\documentclass[journal]{IEEEtran}
\usepackage{amsmath,amssymb,amsfonts}
\usepackage{algorithmic}
\usepackage{algorithm}
\usepackage{array}
\usepackage[caption=false,font=normalsize,labelfont=sf,textfont=sf]{subfig}
\usepackage{textcomp}
\usepackage{stfloats}
\usepackage{url}
\usepackage{verbatim}
\usepackage{multirow}
\usepackage{cite}
\usepackage{graphicx}
\usepackage{picinpar}
\usepackage{url}
\usepackage{flushend}
\usepackage[latin1]{inputenc}
\usepackage{colortbl}
\usepackage{soul}
\usepackage{pifont}
\usepackage{color}
\usepackage{alltt}
\usepackage[colorlinks]{hyperref}
\usepackage{enumerate}
\usepackage{siunitx}
\usepackage{breakurl}
\usepackage{epstopdf}
\usepackage{pbox}
\usepackage{amssymb}
\usepackage{threeparttable}
\usepackage{booktabs}
\usepackage[table]{xcolor}
\usepackage{cuted}
\usepackage{capt-of} 

\begin{document}

\title{Scalable Unseen Objects 6-DoF Absolute Pose Estimation with Robotic Integration}

\author{Jian Liu, Wei Sun, Kai Zeng, Jin Zheng, Hui Yang, Hossein Rahmani, Ajmal Mian, and Lin Wang
\thanks{This work is supported by the National Natural Science Foundation of China under Grants 62473141 and U22A2059, National Science and Technology Major Project of China under Grants 2026ZD1610900, Natural Science Foundation of Hunan Province under Grant 2024JJ5098, Open Foundation of the State Key Laboratory of Advanced Design and Manufacturing for Vehicle Body, and Open Foundation of the Engineering Research Center of Multi-Mode Control Technology and Application for Intelligent System of the Ministry of Education. Ajmal Mian was supported by the Australian Research Council Future Fellowship Award funded by the Australian Government under Project FT210100268. (\emph{Corresponding authors: Wei Sun; Hui Yang.})}
\thanks{Jian Liu, Wei Sun, Kai Zeng, and Hui Yang are with the National Engineering Research Center for Robot Visual Perception and Control Technology, School of Artificial Intelligence and Robotics, Hunan University, Changsha, 410082, China. (e-mail: \{jianliu, wei\_sun, huiyang\}@hnu.edu.cn).}
\thanks{Jin Zheng is with the School of Architecture and Art, Central South University, Changsha, 410082, China.}
\thanks{Hossein Rahmani is with the School of Computing and Communications, Lancaster University, LA1 4YW, United Kingdom.}
\thanks{Ajmal Mian is with the Department of Computer Science and Software Engineering, The University of Western Australia, WA 6009, Australia.}
\thanks{Jian Liu and Lin Wang are with the School of Electrical and Electronic Engineering, Nanyang Technological University, 639798, Singapore.}
}

\maketitle

\begin{strip}
\centering
\vspace{-14em}
\includegraphics[width=\textwidth]{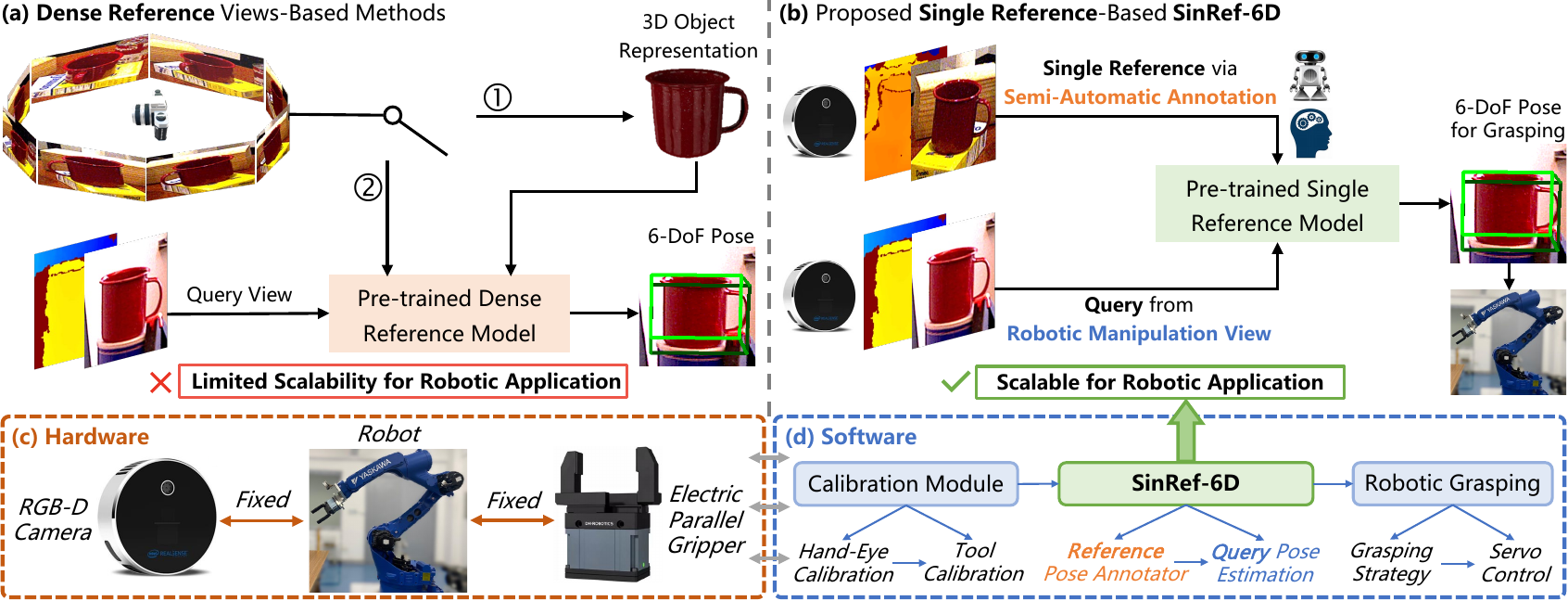}
\captionof{figure}{Overview of the proposed task setup and robotic integration for unseen object 6-DoF absolute pose estimation tailored for practical robotic applications. \textbf{(a)} and \textbf{(b)} compare two types of manual reference view-based methods. \textbf{(a)} Dense reference views-based methods typically rely on {\textcircled{\scriptsize{1}}}: 3D object reconstruction or {\textcircled{\scriptsize{2}}}: template matching, which is time- and memory-consuming (\emph{not suitable for robotic applications}). \textbf{(b)} The proposed method estimates unseen object pose from only a single reference view, providing enhanced efficiency and scalability (\emph{suitable for robotic applications}). \textbf{(c)} and \textbf{(d)} are the detailed hardware and software architectures (see Sec.~\ref{Developed Robotic Grasping System} for further description) of our integrated robotic system. We also develop an efficient semi-automatic annotator based on the proposed task setup, enabling single reference annotation for unseen object within one minute.}
\label{Fig1}
\end{strip}






\markboth{IEEE Transactions on Robotics, ~2026}
{Shell \MakeLowercase{\textit{et al.}}: A Sample Article Using IEEEtran.cls for IEEE Journals}



\begin{abstract}
Pose estimation-guided unseen object 6-DoF robotic manipulation is a key task in robotics. However, the scalability of current pose estimation methods to unseen objects remains a fundamental challenge, as they generally rely on CAD models or dense reference views of unseen objects, which are difficult to acquire, ultimately limit their scalability. In this paper, we introduce a novel task setup, referred to as SinRef-6D, which addresses 6-DoF absolute pose estimation for unseen objects using only a single pose-labeled reference RGB-D image captured during robotic manipulation. This setup is more scalable yet technically nontrivial due to large pose discrepancies and the limited geometric and spatial information contained in a single view. To address these issues, our key idea is to iteratively establish point-wise alignment in a common coordinate system with state space models (SSMs) as backbones. Specifically, to handle large pose discrepancies, we introduce an iterative object-space point-wise alignment strategy. Then, Point and RGB SSMs are proposed to capture long-range spatial dependencies from a single view, offering superior spatial modeling capability with linear complexity. Once pre-trained on synthetic data, SinRef-6D can estimate the 6-DoF absolute pose of an unseen object using only a single reference view. With the estimated pose, we further develop a hardware-software robotic system and integrate the proposed SinRef-6D into it in real-world settings. Extensive experiments on six benchmarks and in diverse real-world scenarios demonstrate that our SinRef-6D offers superior scalability. Additional robotic grasping experiments further validate the effectiveness of the developed robotic system. The code and robotic demos are available at our \href{https://paperreview99.github.io/SinRef-6DoF-Robotic}{homepage}.

\end{abstract}

\begin{IEEEkeywords}
Pose estimation, 6-DoF robotic manipulation, unseen object, single reference, state space model.
\end{IEEEkeywords}

\section{Introduction}
\IEEEPARstart{U}{nseen} object 6-DoF robotic manipulation is a fundamental task underlying scalable robotic applications across diverse domains \cite{tro2024_survey, tro2025, liu2024survey, chen2025tro}. At its core lies the challenge of estimating the 6-DoF absolute pose of 
objects not encountered during training \cite{catre, chen2022sim, fu20226d, Diff9D, tyree2022hope, cyber2025}. Typically, 6-DoF pose comprises of 3-DoF rotation and 3-DoF translation of an object coordinate system relative to the camera coordinate system \cite{NOCS, li2022dcl, cao2022dgecn, wu2022vote, zhou2023deep, ominnocs, omni6d}.

\par 
Object pose estimation methods can be divided into three main categories. \emph{Instance-level methods} \cite{pvnet, se(3)tracknet, self6d, wang2021gdr, di2021so, xu20246d, ffb6d, Dang2024Match, deng2021poserbpf} have attained high precision but are limited to objects encountered during training. In contrast, \emph{category-level methods} \cite{mh6d, ist-net, gpv-pose, ttacope, georef, HouseCat6D, monodiff9d, shi_tro2023} can generalize to objects within the same category but still necessitate retraining for novel object categories. Furthermore, some \emph{unseen object pose estimation methods} \cite{nguyen2022templates, shugurov2022osop, gou2022unseen, hagelskjaer2023keymatchnet, fan2024pope, zhao2024locposenet, pan2024learning} have been proposed recently that {\em do not require retraining} for novel object categories, 
thereby exhibiting enhanced scalability. 


\par Unseen object pose estimation methods can be further divided into two categories: \emph{CAD model-based} \cite{pitteri2019cornet, pitteri20203d, sundermeyer2020multi, okorn2021zephyr}, where a textured CAD model of the unseen object is required during training and inference; \emph{manual reference view-based} \cite{wu2021unseen, corsetti2024open, bundlesdf}, where a set of manually labeled reference views of the unseen object are required. 
Accurate textured CAD models can only be obtained with {specialized equipment and expert knowledge}, which hinders scalability in mobile devices \cite{he_tro2023, talak_tro2023}. Since manual reference views are relatively easy to acquire, methods in this category offer greater scalability. Manual reference view-based methods typically solve pose through 3D object reconstruction or directly obtain coarse pose via template matching, as shown in Fig.~\ref{Fig1} (a), where the switch indicates whether 3D reconstruction is required based on dense reference views. 
Dense reference views consume 
time for acquisition and memory for storage. Furthermore, template matching-based methods require the use of novel template generation techniques or an additional pose refinement 
overhead which further increases the computational complexity.

\par To address the aforementioned challenges, our motivation is to explore a CAD model-free, sparse reference view-based unseen object 6-DoF absolute pose estimation framework, eliminating the need for either 3D object reconstruction or template-based retrieval. Specifically, we formulate the task as the extreme case of sparse reference view, where {\em only a single reference is available}. Motivated by robotic manipulation scenarios, we design a {\em scalable label collection pipeline} where each unseen object is annotated with a single RGB-D reference view in a semi-automatic manner, while absolute pose recovery is obtained through the annotated reference view. The overview of our task setup is shown in Fig.~\ref{Fig1} (b). The robot first captures the object from its default manipulation viewpoint, and a custom-developed annotator provides the corresponding 6-DoF pose label. Given only a single annotated view as the sparse reference prior of an unseen object, our goal is to accurately estimate its 6-DoF absolute pose from arbitrary novel viewpoints in different scenes. However, using a single reference introduces several unique challenges, including large pose discrepancies and limited spatial information.

\par With the task setup, we propose a scalable SinRef-6D framework. \textit{Our key idea is to iteratively establish point-wise alignment between the single reference view and a query view in a common coordinate system to solve the 6-DoF pose of unseen objects.} SinRef-6D introduces two key components: 
1) Iterative object-space point-wise alignment, which addresses large pose discrepancies by leveraging geometric and spatial consistency to refine pose estimation; 2) State Space Models (SSMs), which efficiently capture long-range spatial dependencies from single-view data, offering linear computational complexity and strong spatial modeling capability. 
Specifically, we propose to align the reference and query point clouds within the object coordinate system (Sec. \ref{Points Focalization}). Given the importance of spatial information for point-wise alignment and the need for a lightweight model for mobile deployment,
we introduce Point and RGB SSMs (Sec. \ref{SSMs}) to establish point-wise alignment for pose solving (Sec. \ref{Point-wise Alignment & Pose Solving}).
To handle the potentially large pose discrepancies between the reference and query views, we propose to iteratively refine the alignment in the object coordinate system, which gives more accurate and robust pose estimation (Sec. \ref{Training Mode}). Furthermore, we develop a complete hardware-software robotic system that integrates the proposed SinRef-6D to evaluate its scalability in real-world scenarios (Sec. \ref{Developed Robotic Grasping System}), as shown in Fig. \ref{Fig1} (c) and (d). Our main contributions are summarized as follows:
\begin{itemize}
    \item We introduce an efficient and scalable task setup for unseen object 6-DoF absolute pose estimation using only a single reference view captured during robotic manipulation, eliminating the need for computation-intensive template matching and multi-view reconstruction. \textcolor{black}{We further develop an integrated hardware-software robotic system tailored to the proposed task setup and framework, validating their efficacy in real-world scenarios.} 
    \item We propose an object-space point-wise alignment strategy with iterative refinement, facilitating direct alignment of query and reference views while effectively handling large pose discrepancies. This enhances geometric consistency and spatial awareness, enabling unseen object pose estimation without category-specific retraining.
    \item We propose Point and RGB SSMs to capture rich spatial information for establishing point-wise alignment, enabling efficient long-range spatial modeling with linear computational complexity. 
    \item Extensive experiments demonstrate that our task setup and framework enable highly scalable 6-DoF robotic grasping of unseen objects in diverse environments.
\end{itemize}

\par The remainder of this paper is structured as follows. Sec. \ref{Related Work} reviews recent advances in unseen object pose estimation. Sec. \ref{Methodology} introduces the proposed task setup and corresponding framework. Sec. \ref{Developed Robotic Grasping System} describes the developed 6-DoF robotic grasping system that integrates both hardware and software. Sec. \ref{Experiments} presents comprehensive experimental results that validate the scalability of SinRef-6D and the effectiveness of the robotic system. Finally, Sec. \ref{Conclusion} summarizes the paper.

\section{Related Work}\label{Related Work}
This section provides an overview of state-of-the-art methods in unseen object absolute (Sec. \ref{CAD Model-based Methods} and Sec. \ref{Manual Reference View-based Methods}) and relative (Sec. \ref{Relative Pose Estimation Methods}) pose estimation, followed by a discussion on how our work differs from existing approaches.

\subsection{CAD Model-based Methods}\label{CAD Model-based Methods}
Research in the domain of CAD model-based methods first require obtaining the precise CAD model of the unseen object, which is then used as prior knowledge for pose estimation. These methods can be further categorized into 1) feature matching-based and 2) template matching-based.

\par \emph{Feature matching-based methods} \cite{chen2023zeropose, GCPose, SAM-6D, FreeZe, MatchU} learn a model to match features between the observed image and CAD model, establishing 2D-3D or 3D-3D correspondences to estimate object pose. Specifically, GCPose \cite{GCPose} proposes a geometry correspondence-based approach that leverages generic, object-agnostic geometric features to establish clear and robust 3D-3D correspondences.
SAM-6D \cite{SAM-6D} introduces a novel matching score based on semantics, appearance, and geometry to improve segmentation. For pose estimation, it employs a two-stage point matching model to establish dense 3D-3D correspondences. FreeZe \cite{FreeZe} develops a method that combines visual and geometric features from various pre-trained models to improve pose prediction stability and accuracy. MatchU \cite{MatchU} proposes a technique for predicting object pose from RGB-D images by integrating 2D texture with 3D geometric cues.

\par \emph{Template matching-based methods} \cite{foundpose, wang2024object, megapose, genflow, gigapose} render multiple template views of the object with different poses from the CAD model. Then, they retrieve the template that best matches the observed image to obtain a coarse pose, followed by a refinement process to achieve accurate pose estimation. For example, MegaPose \cite{megapose} proposes a render-and-compare-based method and a coarse-to-fine pose estimation strategy. GenFlow \cite{genflow} introduces a shape-constrained recurrent flow framework that predicts optical flow between the query and template images while iteratively refining the pose. GigaPose \cite{gigapose} achieves fast and robust pose estimation by striking an effective balance between template matching and patch correspondences. FoundationPose \cite{foundationpose} increases the quantity and diversity of synthetic data based on diffusion model and achieves superior performance through render-and-compare.

\subsection{Manual Reference View-based Methods}\label{Manual Reference View-based Methods}
To eliminate the need for a precise CAD model, manual reference view-based methods employ manual reference views as the prior knowledge for unseen objects. These methods can also be categorized into 1) feature matching-based and 2) template matching-based.

\par \emph{Feature matching-based methods} \cite{fs6d, onepose, onepose++, castro2023posematcher, lee2024mfos} aim to establish 3D-3D correspondences between the query view and reference views, or 2D-3D correspondences between the query view and the 3D object representation reconstructed from reference views. Specifically,  FS6D \cite{fs6d} proposes a dense prototype matching method to explore geometric and semantic relations between the query view and reference views, estimating the pose of unseen objects using only a few reference views. OnePose \cite{onepose} first utilizes Structure from Motion (SfM) to reconstruct the 3D representation of the unseen object using all reference views, and then establishes 2D-3D correspondences between the query view and the reconstructed 3D representation using a graph attention network. OnePose++ \cite{onepose++} introduces a keypoint-free SfM method to reconstruct a semi-dense 3D representation of textureless objects by leveraging the detector-free feature matching approach LoFTR \cite{sun2021loftr}, enhancing robustness against textureless objects.

\par \emph{Template matching-based methods} \cite{latentfusion, gen6d, cai2024gs, sa6d, du2022pizza} primarily utilize a retrieval and refinement strategy. They directly use labeled reference views as templates to retrieve a coarse pose, followed by a refinement process to enhance accuracy. Specifically,  LatentFusion \cite{latentfusion} reconstructs 3D object representation and estimates translation using bounding boxes and depth values. Then, the initial rotation is determined by angle sampling and further refined through gradient updates using render and compare. Gen6D \cite{gen6d} first detects object bounding boxes, then compares the query and reference images via similarity scores to obtain an initial pose. Next, the pose is refined via a proposed refiner. FoundationPose \cite{foundationpose} introduces an object-centric neural field to enable accurate 3D object modeling and RGB-D rendering, achieving performance comparable to instance-level methods. GS-Pose \cite{cai2024gs} joints segmentation and introduces a 3D gaussian splatting-based refiner, which simultaneously enhances the accuracy of object localization and pose estimation.

\subsection{Unseen Object Relative Pose Estimation Methods}\label{Relative Pose Estimation Methods}
Relative object pose estimation \cite{3dahv, dvmnet, nope, unopose, one2any, any6d} refers to computing the pose transformation of an object between two different views. 3DAHV \cite{3dahv} proposes a 3D-aware hypothesis-and-verification framework for relative pose estimation of unseen objects from a reference image, achieving robust generalization under large pose variations without relying on dense multi-view supervision. Building on this idea, DVMNet \cite{dvmnet} introduces an end-to-end voxel-based framework that bypasses discrete hypothesis generation by directly aligning voxelized 3D features from two RGB images, resulting in improved accuracy and reduced computational cost. In contrast, NOPE \cite{nope} presents a fast, training-free method that estimates relative pose by predicting pose-conditioned viewpoint embeddings using an attention-enhanced U-Net, without requiring 3D models. While these methods demonstrate strong scalability, the absence of depth information limits their ability to estimate the full 3-DoF relative translation.

\textcolor{black}{More recently, some works \cite{unopose, one2any, any6d} have explored pose estimation using a single RGB-D reference view to reduce onboarding cost for unseen objects. UNOPose \cite{unopose} incorporates depth data and proposes a one-reference-based pose estimation framework that constructs an SE(3)-invariant reference representation and adaptively weights correspondences to handle low viewpoint overlap. One2Any \cite{one2any} further introduces a category-agnostic method for 6-DoF object pose estimation that leverages a reference-query RGB-D pair to generate pose embeddings and decode object coordinates. Any6D \cite{any6d} estimates both object pose and size from an RGB-D anchor image by leveraging joint object alignment and a render-and-compare strategy. Despite their effectiveness, these methods primarily focus on relative pose estimation between the reference and query views, which is insufficient for robotic manipulation scenarios where absolute object poses in a common coordinate system are required for action execution. In contrast, our work targets single-reference 6-DoF absolute pose estimation under robotic manipulation settings. To this end, we introduce a semi-automated reference acquisition and annotation pipeline, a single reference view-based point cloud focalization strategy to establish a common coordinate system, and SSMs-based feature extraction networks tailored for the limited geometric and spatial information available from a single view. This problem-driven design enables direct deployment in manipulation pipelines while maintaining scalability to unseen objects.}

\vspace{0.5em}
\noindent \textbf{Discussions:} Overall, CAD model-based methods depend on textured CAD models, and manual reference view-based methods require dense reference views, both adding manual effort in real-world applications. \textcolor{black}{Related works such as FoundationPose \cite{foundationpose} also employ transformer-based architectures for iterative pose refinement; however, our SSM-based backbone is explicitly designed to model long-range spatial dependencies under severely limited geometric information, which is particularly critical in our single-reference setting.} Additionally, relative pose estimation methods are not well-suited for robotic manipulation tasks that require absolute poses for action execution. Hence, this paper seeks to enable unseen object 6-DoF absolute pose estimation with a single reference view, reducing manual overhead and enhancing scalability for robotic applications. Most recently, 3D foundation models such as SAM 3D \cite{chen2025sam-3d} and VGGT \cite{wang2025vggt} suggest a clear trend toward large-scale, data-driven geometric perception. However, these advances do not diminish the importance of reliable pose estimation; instead, they increase the demand for scalable modules that can provide accurate geometric initialization for annotation bootstrapping and downstream reasoning. In this broader context, our method can also be viewed as a complementary component: a practical and scalable solution for unseen object 6-DoF pose estimation that remains valuable even as 3D foundation models continue to evolve.

\section{Methodology}\label{Methodology}
We begin with an overview of the overall task setup and framework, including its input and output (Sec. \ref{Overview}). We then describe the initialization process, which involves unseen object segmentation from the input RGB-D image (Sec. \ref{Initialization}). Next, we present the proposed point focalization strategy (Sec. \ref{Points Focalization}), Point and RGB SSMs (Sec. \ref{SSMs}), and point-wise alignment for pose solving (Sec. \ref{Point-wise Alignment & Pose Solving}). Finally, we explain the training procedure and supervision scheme (Sec.~\ref{Training Mode}).

\subsection{Task Setup and Framework Overview}\label{Overview}
\par \textcolor{black}{Overall, our work is problem-driven, aiming to enable scalable  6-DoF absolute pose estimation of unseen objects for robotic manipulation with minimal prior information. To this end, we present a unified system that operates with only a single reference view, where each component is explicitly designed to address the challenges arising from single-reference, manipulation-oriented absolute pose estimation.} Specifically, for unseen objects which are not encountered during training, SinRef-6D takes a pair of RGB-D images captured from robotic manipulation viewpoints as \emph{input}: a single reference image and a query image. The reference image is selected only once and annotated with a 6-DoF pose through a semi-automatic manner using our custom-developed pose annotator. The \emph{output} of SinRef-6D is the estimated 6-DoF absolute pose of the unseen object in the arbitrary query image. Algorithm \ref{algo:sinref6d} shows the overall pipeline of the proposed framework.


\begin{algorithm}[t!]
\renewcommand{\baselinestretch}{1.15}
\caption{{\fontsize{10pt}{11pt}\selectfont Overall Pipeline of SinRef-6D}}
\label{algo:sinref6d}
\begin{algorithmic}[1]
{\fontsize{10pt}{11pt}\selectfont
\STATE \textbf{Input:} Reference RGB-D image; Query RGB-D image
\STATE \textbf{Output:} Estimated 6-DoF object pose $[R_{final} \mid t_{final}]$
\STATE Segment reference and query images to obtain object RGB masks $I_r$, $I_q$ and corresponding depth masks via SAM-driven similarity matching
\STATE Back-project depth masks to get object reference and query point clouds $P_r, P_q$
\STATE Transform $P_r$ into the object coordinate system using known reference pose (obtained via a semi-automated annotator) $[R_r \mid t_r]$: $P_r^o = R_r^\top (P_r - t_r)$
\STATE Initialize query pose:\\$[R_1 \mid t_1] \gets \text{identity matrix}, \text{average coordinate} (P_q)$
\FOR{$i = 1$ to $T$} 
    \STATE Transform $P_q$ into the object coordinate system:\\$P_q^i = R_i^\top (P_q - t_i)$
    \STATE Extract point-wise and RGB features:
        \STATE \hspace{0.5cm} $F_r \gets {Point} \text{ } {SSM}(P_r^o) \oplus {RGB} \text{ } {SSM}(I_r)$
        \STATE \hspace{0.5cm} $F_q^i \gets {Point} \text{ } {SSM}(P_q^i) \oplus {RGB} \text{ } {SSM}(I_q)$
    \STATE Perform point-wise feature alignment:
        \STATE \hspace{0.5cm} $\bar{F}_r, \bar{F}_q^i \gets \text{GeoTransformer}(F_r, F_q^i)$
        \STATE Compute point-wise affinity: $A^i = \bar{F}_q^i \otimes \bar{F}_r^\top$
    \STATE Estimate 6-DoF object pose via weighted SVD:
        \STATE \hspace{0.5cm} $[R_{i+1}, t_{i+1}] = {WSVD}(A^i, P_r^o, P_q)$
\ENDFOR
\STATE \textbf{Return:} $[R_{final}, t_{final}] \gets [R_{T+1}, t_{T+1}]$
}
\end{algorithmic}
\end{algorithm}

\par
\begin{figure*}[t!]
    \centering
    \includegraphics[width=\textwidth]{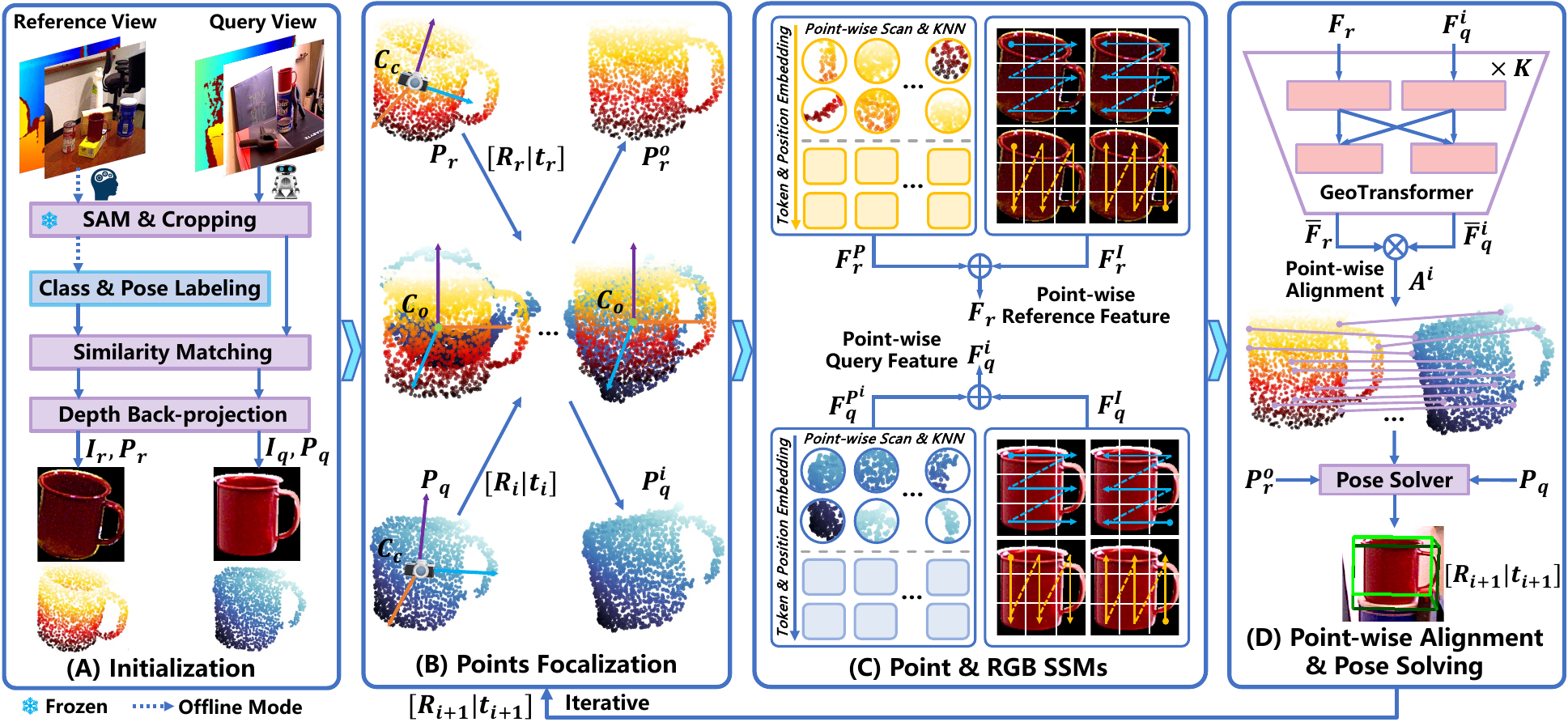}
    \caption{Our proposed SinRef-6D framework. Given a normal RGB-D reference view of an unseen object, we aim to predict its 6-DoF absolute pose from any query view. SinRef-6D comprises four modules: (A) The reference view is labeled via a semi-automatic annotator, then the RGB-D images of the reference and query views are segmented, and the segmented depth maps are back-projected into point clouds. (B) The corresponding point clouds of the reference and query views are focalized from the camera coordinate system to the object coordinate system. (C) Leveraging the proposed Point and RGB SSMs (details are shown in Fig. \ref{Fig.PSS} and Fig. \ref{Fig.VSS}), features are extracted from the focalized point clouds and RGB images, forming point-wise reference and query features. (D) These features are then used to establish point-wise alignment to solve the object pose. Finally, the computed pose is fed back into module (B) to iteratively improve the accuracy of the point-wise alignment, yielding a more precise object pose.}
    \label{framework}
\end{figure*}

 Figure~\ref{framework} shows the overall workflow that comprises four main components: (A) \emph{Initialization} segments the unseen object in the input reference and query views. (B) \emph{Points Focalization} focalizes the unseen object in the reference and query views into the object coordinate system using their corresponding poses. (C) \emph{Point \& RGB SSMs} employ state space models to extract point-wise reference and query features. (D) \emph{Point-wise Alignment \& Pose Solving} derives point-wise alignment relationships using the features extracted in (C) to solve the object pose in the query view. In addition, iterating the process from (B) to (D) allows for further obtaining more accurate point-wise alignment and object pose.

\subsection{Initialization}\label{Initialization}
\textcolor{black}{Notably, randomly sampling arbitrary rendering viewpoints may introduce extreme perspectives (e.g., near top-down views) that deviate significantly from real-world reference acquisition, whereas manually selecting viewpoints for all objects would be labor-intensive and may reduce robustness to viewpoint variations. From a practical real-world application perspective, during training, the synthetic reference view is sampled from a viewpoint range that approximates the robotic manipulation viewpoint while introducing natural perturbations.} Importantly, this reference view is \emph{not carefully selected}. To support this claim, we additionally generate reference views using the same rendering protocol as GigaPose \cite{gigapose}. Specifically, we randomly sample one viewpoint from the \emph{50th} to the \emph{120th} in its rendering sequence, which is designed to approximate the robotic manipulation viewpoint. This simulates manual reference view acquisition while introducing natural pose perturbations. During the evaluation in real-world robotic scenarios, we adopt a semi-automatic manner. 
The reference view for each unseen object is captured by the robot from an occlusion-free manipulation viewpoint and annotated using our custom-developed annotator. The rotation is determined using a calibration board, while the translation and size are manually adjusted through keyboard control (some visualizations are shown in the first row of Fig.~\ref{graspsence}). 
For testing on public benchmarks, we adopt both reference view acquisition strategies to align with those used in training.

\par The pipeline of the initialization process is shown in part (A) of Fig. \ref{framework}. Since both the reference and query views often contain cluttered backgrounds, we first segment the background. For a fair comparison, we employ Mask R-CNN \cite{maskrcnn} or zero-shot CNOS \cite{cnos} with FastSAM to segment the input images, and then back-project the segmented depth maps into point clouds. This results in the segmented RGB images and point clouds for both reference (${I_r, P_r \in \mathbb{R}^{{N_r} \times 3}}$) and query (${I_q, P_q \in \mathbb{R}^{{N_q} \times 3}}$) views, where ${N_r}$ and ${N_q}$ denote the number of points in the reference and query point clouds, respectively. Notably, CNOS relies on object CAD models for rendering template images, which contrasts with our CAD model-free setup. Based on this, we also use only our single reference view as the template image for similarity matching in CNOS segmentation (see the first two rows of Tab.~\ref{label2} for details)\cite{cnos}.

\subsection{Points Focalization}\label{Points Focalization}
Since SinRef-6D aims to iteratively align point clouds for precise object pose solving, our first step is to focalize the reference and query point clouds within a common coordinate system.
This focalization facilitates point-wise alignment, ensures geometric consistency during iterative refinement, and inherently decouples pose estimation from category priors, enhancing robustness to unseen objects. Specifically, as the reference point cloud ${P_r}$ has a pose annotation $\left[ {{R_r}|{t_r}} \right]$, we can transform it from the camera coordinate system ${C_c}$ to the object coordinate system ${C_o}$ as follows:
\begin{equation} \label{equation1}
P_r^o = {R_r^\top}\left( {{P_r} - {t_r}} \right),
\end{equation}
where ${t_r}$ and ${R_r}$ denote the annotated translation and rotation, respectively. $\top$ denotes matrix transpose, ${P_r^o}$ denotes the reference point cloud in the object coordinate system.

\par For the query point cloud, we apply the same method to transform it into the object coordinate system as follows:
\begin{equation} \label{equation2}
P_q^i = {R_i^\top}\left( {{P_q} - {t_i}} \right),
\end{equation}
where ${t_{i}}$ and ${R_{i}}$ represent the translation and rotation of the object in the $i$\emph{-th} iteration. ${P_q^i}$ represents the query point cloud in the object coordinate system after the $i$\emph{-th} iteration. Since the object pose in the query view is initially unknown, we do not perform rotation transformation during the first points focalization and instead set the translation ${t_1}$ to the average coordinate of the object. In subsequent iterations, we use the object pose $\left[ {{R_{i+1}}|{t_{i+1}}} \right]$ solved in the previous round for coordinate transformation. The overall process is shown in part (B) of Fig.~\ref{framework}.

\subsection{Point \& RGB SSMs}\label{SSMs}
\begin{figure}[t!]
    \centering
    \includegraphics[width=\linewidth]{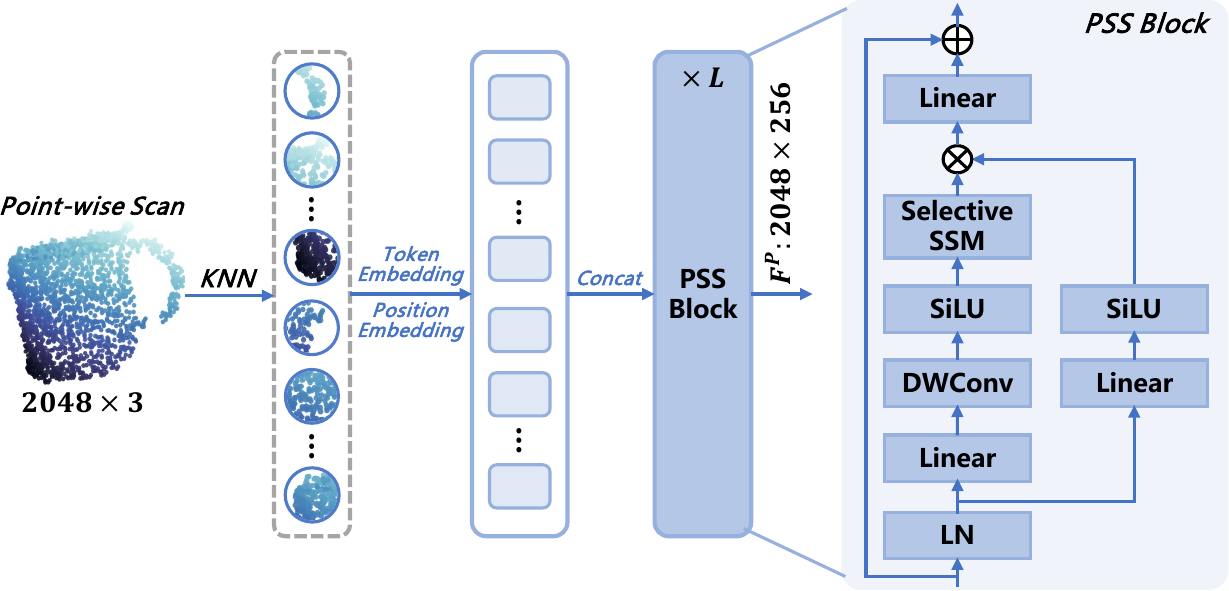}
    \vspace{-1.25em}
    \caption{Architecture of the proposed Point SSM. It takes object point clouds as input and captures point-wise features with fine-grained spatial semantics.}
    \label{Fig.PSS}
    \vspace{-0.25em}
\end{figure}

\begin{figure*}[t!]
    \centering
    \includegraphics[width=0.92\linewidth]{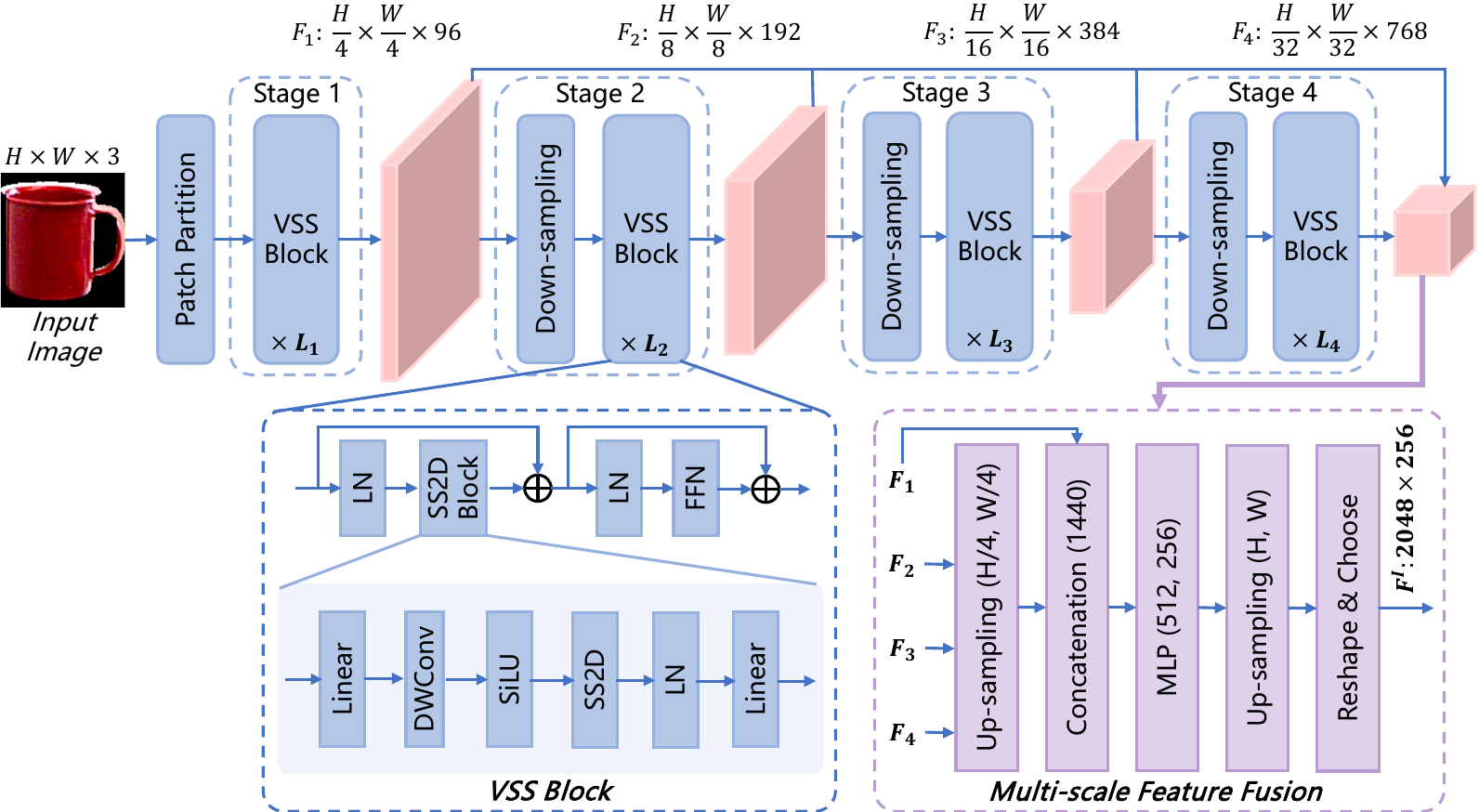}
    \caption{Detailed architecture of the proposed RGB SSM. It takes segmented RGB mask images as input, where $H$ and $W$ denote the height and width of the image, and outputs rich visual features for subsequent reasoning.
    }
    \label{Fig.VSS}
\end{figure*}

Since point-wise alignment relies on rich spatial features, sequential modeling of point clouds and RGB images enables effective long-range spatial encoding, enhancing feature discrimination and geometric consistency for more precise alignment. To handle limited spatial cues and real-time demands, we adopt a simple-yet-efficient design, incorporating lightweight Point and multiscale RGB SSMs for efficient long-range modeling from sparse single-view data with linear complexity.
The selective scan structured state space sequence (S6) models \cite{S6model} represent a class of sequence models that excel in sequence handling. These models extend the earlier S4 model \cite{S4model}, mapping an input sequence $x\left( t \right) \in \mathbb{R} \to y\left( t \right) \in \mathbb{R}$ through a latent state $h\left( t \right) \in {\mathbb{R}^M}$ according to the ordinary linear differential equations:
\begin{equation} \label{equation3}
\begin{array}{l}
{h^{\prime}}\left( t \right) = \emph{\textbf{A}}h\left( t \right) + \emph{\textbf{B}}x\left( t \right),\\
y\left( t \right) = \emph{\textbf{C}}h\left( t \right) + \emph{\textbf{D}}x\left( t \right),
\end{array}
\end{equation}
where $\emph{\textbf{A}} \in {\mathbb{R}^{M \times M}},\emph{\textbf{B}} \in {\mathbb{R}^{M \times 1}},\emph{\textbf{C}} \in {\mathbb{R}^{1 \times M}},$ and $\emph{\textbf{D}} \in {\mathbb{R}^{1}}$ are weighting parameters. Specifically, the continuous dynamical systems are discretized using the following zero-order hold discretization method in practical computations:
\begin{equation} \label{equation4}
\begin{array}{l}
{h_t} = \bar {\emph{\textbf{A}}}{h_{t - 1}} + \bar {\emph{\textbf{B}}}{x_t},~~{y_t} = {\emph{\textbf{C}}}{h\left( t \right)},\\
\bar {\emph{\textbf{A}}} = \exp \left( {\Delta {\emph{\textbf{A}}}} \right),\bar {\emph{\textbf{B}}} = {\left( {\Delta {\emph{\textbf{A}}}} \right)^{ - 1}}\left( {\exp \left( {\Delta {\emph{\textbf{A}}}} \right) - {\emph{\textbf{I}}}} \right) \cdot \Delta {\emph{\textbf{B}}},
\end{array}
\end{equation}
where $\Delta$ denotes the discrete step size. Given that both the weighting parameters and discretization method remain constant over time, S4 models can be considered linear time-invariant systems. S6 model \cite{S6model} further extends the projection matrices of S4 models to enable a selective scan of the entire input sequence.
Specifically, we model the sequences of point clouds and RGB images by designing a Point SSM (see Fig. \ref{Fig.PSS}) and an RGB SSM (see Fig. \ref{Fig.VSS}) as follows:
\begin{equation} \label{equation5}
\begin{array}{l}
{F_r} = Point\ SSM\left( {P_r^o} \right) \oplus RGB\ SSM\left( {{I_r}} \right),\\
F_q^i = Point\ SSM\left( {P_q^i} \right) \oplus RGB\ SSM\left( {{I_q}} \right),
\end{array}
\end{equation}
where $ \oplus $ represents matrix addition. ${F_r}\in {\mathbb{R}^{{N_r} \times C}}$ and ${F_q^i\in {\mathbb{R}^{{N_q} \times C}}}$ represent the point-wise reference features and the point-wise query features at the \emph{i-th} iteration, respectively. $C$ is the dimension of feature channels.

\par \textcolor{black}{The sequence modeled by the SSMs refers to an ordered collection of spatial tokens rather than physical time steps. For the RGB branch, the input image is partitioned into patches and flattened into sequences following two fixed raster-scan orders (as shown in part (C) of Fig. \ref{framework}), which are consistently used during training and inference. This ordering enables the SSM to capture long-range spatial dependencies across image regions without encoding temporal information. For the point cloud branch, due to the intrinsic unordered nature of point sets, the input sequence is constructed by iterating over all points, and the resulting order is neither fixed nor semantically meaningful. For the Point SSM, as shown in Fig. \ref{Fig.PSS}, we first perform a point-wise scan and use K-Nearest Neighbor (KNN) to sample a set of points for each scanned point to form a token. Then, we compute all token embeddings and add a position embedding to them. Subsequently, the token embeddings are concatenated and passed into the points state space (PSS) blocks to obtain the point-wise feature ${F^P}\in{^{2048 \times 256}}$. The details of the selective SSM in PSS blocks can be found in the S6 model \cite{S6model}. For RGB image feature extraction, we propose an RGB SSM based on the cross-scan manner and multi-scale feature fusion, as depicted in Fig.~\ref{Fig.VSS}. The architecture consists of four stages, where each stage employs visual state space (VSS) blocks \cite{Vmamba} to extract image features at different scales. These multi-scale features are then fused, reshaped, and chosen by using the image mask to obtain the final image feature representation ${F^I}\in{^{2048 \times 256}}$.}

\par \textcolor{black}{Specifically, $F^P$ and $F^I$ denote the generic point-wise and image feature representations, which are instantiated as reference and query features. The complete process is illustrated in part (C) of Fig.~\ref{framework}, where $F_r^P$ and $F_q^{{P^i}}$ represent the extracted point-wise reference and query features (at the \emph{i-th} iteration), while $F_r^I$ and $F_q^I$ denote the extracted features from the reference and query RGB images.}

\subsection{Point-wise Alignment \& Pose Solving}\label{Point-wise Alignment & Pose Solving}
Upon acquiring the point-wise reference and query features, our objective is to develop a model with the scalability to handle unseen objects to establish point-wise alignment. This model provides enhanced learnability compared to the direct pose regression model for unseen objects. Specifically, we input ${F_r}$ and ${F_q^i}$ into the GeoTransformer \cite{GeoTransformer}, where they undergo geometric-aware self-attention and cross-attention, yielding the final point-wise reference and query features ${\bar F_r}$ and ${\bar F_q^i}$. Then, we obtain the point-wise affinity matrix ${A^i}$ as follows and select the point pairs with the highest similarity for alignment:
\begin{equation} \label{equation6}
{A^i} = \bar F_q^i \otimes {\bar F_r}^\top,
\end{equation}
where $ \otimes $ represents matrix multiplication.

\par The point-wise alignment of the reference and query view point clouds in the object coordinate system is visualized in part (D) of Fig.~\ref{framework}. Once the point-wise alignment relationship is established, we can directly solve the 6-DoF object pose using the weighted singular value decomposition (WSVD) algorithm as follows:
\begin{equation} \label{equation7}
\left[ {{R_{i + 1}}|{t_{i + 1}}} \right] = {WSVD}\left( {{A^i},P_r^o,{P_q}} \right),
\end{equation}
where rotation ${R_{i + 1}}$ and translation ${t_{i + 1}}$ denote the 6-DoF object pose solved from the \emph{i-th} iteration.

\par Given the considerable pose discrepancies between the initial query and reference point clouds (especially in terms of rotation), misaligned point pairs may occur during the alignment process, which will result in inaccurate pose estimation. To mitigate this issue, we introduce an iterative alignment strategy that iteratively performs steps (B) to (D) outlined in Fig.~\ref{framework}. Specifically, the estimated object pose from (D) is fed back into (B) to iterate the focalization of the query point cloud. This iterative strategy facilitates gradual convergence of the query and reference point clouds, ultimately yielding more precise object pose estimation.

\subsection{Training Mode}\label{Training Mode}
During practical deployment, we observe that the initial query and reference point clouds have significant pose discrepancies. Subsequently, their pose differences become smaller after each iteration. Using a single GeoTransformer with shared weights during the iterative process (D) in Fig.~\ref{framework} could adversely affect the accuracy of object pose estimation. Even multi-view transformers have shown that transformer decoder layers can implicitly implement an iterative refinement process for geometric estimation. For example, Stary \emph{et al.} \cite{stary2025understanding} analyzed multi-view transformer architectures and demonstrated that the internal feature state evolves across decoder blocks, progressively refining correspondences and relative camera pose estimates. In such settings, a single network can effectively perform both coarse pose estimation and fine refinement within one unified architecture. However, unseen object pose estimation presents a different regime. Unlike multi-view camera pose estimation, which benefits from strong multi-view geometric constraints and global scene consistency, our setting must handle novel object geometries and limited observations from a single reference view. These factors introduce substantial ambiguity during the initial alignment stage, which we found difficult to reconcile with the objectives of fine-grained pose refinement when using a single backbone.

\textcolor{black}{Motivated by the above observation, we employ two GeoTransformer models with unshared weights to explicitly handle different alignment regimes. The first model is responsible for the initial point-wise alignment, where the pose discrepancy between the reference and query views is typically large and corresponds to a coarse alignment problem. After this stage, the pose discrepancy is significantly reduced, and a second GeoTransformer is used for subsequent iterative refinement under a local alignment regime. This separation allows each model to specialize in a distinct input distribution, improving stability and accuracy during refinement.} Detailed rationale for this choice is given in Sec.~\ref{Ablation Study} and Tab.~\ref{label4}.

We employ two GeoTransformer models with unshared weights to explicitly handle different alignment regimes. The first model is responsible for the initial point-wise alignment, where the pose discrepancy between the reference and query views is typically large and corresponds to a coarse alignment problem. After this stage, the pose discrepancy is significantly reduced, and a second GeoTransformer is used for subsequent iterative refinement under a local alignment regime. This separation allows each model to specialize in a distinct input distribution, improving stability and accuracy during refinement.


\par Since the object pose is solved from the point-wise alignment relationships, we supervise this alignment using the following cross-entropy loss during training:
\begin{equation} \label{equation8}
\textcolor{black}{Loss = \sum\limits_{i = 1,...,K} {\left( {CE( {{A^i},{{\bar p}_q}} ) + CE( {{A^{{i^{\top}}}},{\bar p}_r} )} \right)},}
\end{equation}
\textcolor{black}{where $K$ denotes the number of point-wise alignment iterations used during training.} $CE\left( { \cdot , \cdot } \right)$ represents the cross-entropy loss function. ${{\bar p}_q} \in {\mathbb{R}^{N_q}}$ and ${{\bar p}_r} \in {\mathbb{R}^{N_r}}$ represent the ground truth for $P_q^i$ and $P_r^o$. Each element ${p_q}$ in ${{\bar p}_q}$, corresponding to the point $p_q^i$ in $P_q^i$, can be simply obtained using the index of the closest point in $P_r^o$ to $p_q^i$ from the given ground-truth rotation ${R_{gt}}$ and translation ${t_{gt}}$. Note that if the nearest point distance exceeds a specified threshold (set at 15 centimeters in this paper), we discard that point pair. In addition, the elements in ${{\bar p}_r}$ are obtained in the same manner.

\section{Developed Robotic Grasping System}\label{Developed Robotic Grasping System}
With the estimated pose, this section presents our developed hardware-software robotic grasping system, which integrates the proposed SinRef-6D into it in real-world settings.

\subsection{Developed Hardware and Software System}
The overall system consists of hardware and software components, with their interconnections illustrated in Fig. \ref{Fig1}. The hardware includes an Intel RealSense L515 RGB-D camera, a Yaskawa MOTOMAN-MH12 robotic arm, and a DH-PGI-140-80 electric parallel gripper. The RGB-D camera and gripper communicate with the software part via USB, while the robot is connected through Gigabit Ethernet. The camera is mounted in an eye-in-hand configuration on the robotic arm.

The software component of our integrated robotic system consists of three main modules: a calibration module, an unseen object pose estimation module, and a robotic grasping module. These modules are integrated through a unified graphical user interface. The calibration module performs both hand-eye and tool calibration, using the HALCON-based method and a five-point approach, respectively. The pose estimation module supports pose annotation for the reference view as well as pose inference for novel query views. We design a semi-automatic CAD-free pose annotator for unseen objects, where object rotation is solved via a calibration board (as shown in the first two rows of Fig. \ref{graspsence}), while translation and size are manually aligned via keyboard control. Overall, our annotator enables 6-DoF pose labeling of a single reference view in \emph{less than one minute} per unseen object. Lastly, the grasping module includes a grasp strategy planner and a servo control interface for executing 6-DoF grasps.


\subsection{6-DoF Robotic Grasping Workflow}
\subsubsection{Overall Workflow}
The complete pipeline for 6-DoF robotic grasping in 3D space can be formulated as:
\begin{equation}\label{equation}
T_{o2t} = T_{e2t} \otimes T_{c2e} \otimes T_{o2c},
\end{equation}
where $ \otimes $ denotes matrix multiplication. $T_{o2t}$ represents the transformation from the object coordinate to the tool coordinate systems, which is the target transformation required for executing a robotic grasp. The term $T_{e2t}$ denotes the transformation from the robot end coordinate to the end-effector (tool) coordinate systems, obtained via tool calibration. $T_{c2e}$ is the transformation from the camera coordinate to the robot end coordinate systems, computed through hand-eye calibration. Finally, $T_{o2c}$ refers to the transformation from the object coordinate to the camera coordinate systems, which is \emph{the most critical component} and is estimated by the proposed unseen object pose estimation method.

\subsubsection{Multi-object Grasping Strategy}
With the estimated 6-DoF object pose, we employ a lightweight grasping strategy to enable continuous multi-object robotic grasping \cite{stg6d}. Specifically, we adopt a depth-based sequential grasping strategy, where objects are ordered by the depth (relative to the camera) of their center points. \textcolor{black}{For each object, the estimated 3-DoF translation determines the target grasp point, \emph{i.e.}, the position to which the end-effector center is moved, the grasping direction is defined by a vector from its closest visible point projected onto the $z$-axis to the center point of the object coordinate system, while the gripper closes along the $x$-axis.} For safe execution, the grasp point is translated upward by 2 cm relative to the tabletop. When the angle between the estimated object $z$-axis and the inward normal of the tabletop falls outside the range of $[20^\circ, 60^\circ]$, the grasp orientation is clamped to $30^\circ$ to ensure stable and collision-free grasping. This simple grasping strategy proves effective for both symmetric and asymmetric objects, despite existing axis ambiguities.

\section{Experiments}\label{Experiments}
We first introduce the benchmarks and evaluation metrics (Sec.~\ref{Datasets and Evaluation Metrics}), followed by the implementation details (Sec.~\ref{Implementation Details}). We then compare SinRef-6D with both manual reference view-based and CAD model-based methods on these real-world benchmarks to validate its superior performance (Sec.~\ref{Quantitative Comparisons with SOTA Methods} and Sec.~\ref{Qualitative Analysis}). Next, we evaluate the effectiveness of our approach in real-world robotic grasping scenarios by deploying it on our integrated hardware-software robotic system to perform grasping tasks (Sec.~\ref{Real-world Robotic Grasping}). Finally, we present comprehensive ablation studies to analyze the contributions of key components, the influence of point cloud alignment iterations, and the effect of random reference view selection (Sec.~\ref{Ablation Study}).

\subsection{Datasets and Evaluation Metrics}\label{Datasets and Evaluation Metrics}
\noindent \textbf{Datasets:} We conduct extensive experiments on six benchmark datasets (LineMod \cite{linemod}, LM-O \cite{linemod-o}, TUD-L \cite{2024bop}, IC-BIN \cite{IC-BIN}, HB \cite{HB}, and YCB-V \cite{posecnn}) and real-world robotic scenes. For a fair comparison, we follow the BOP Challenge setting \cite{2024bop} to train on the synthetic dataset generated by MegaPose \cite{megapose} using the ShapeNet-Objects \cite{shapenet} and Google-Scanned-Objects \cite{googlescan} datasets. This training dataset comprises $ \sim $2 million images from $ \sim $50K objects.

\begin{table*}[!t]
\renewcommand{\arraystretch}{1.5}
\centering
\caption{Comparison of SinRef-6D with other manual reference view-based methods on the LineMod dataset \cite{linemod}, evaluated using the ADD-0.1d metric. ``Ref." and ``Recon." mean "Reference" and "Reconstruction". $^\dag$ represents Gen6D \cite{gen6d} without fine-tuning. $^\wedge$ indicates that the reference view is manually selected from the corresponding dataset to approximate the robotic manipulation viewpoint during both training and testing.}
\resizebox{\linewidth}{!}{
\begin{tabular}{l|ccc|ccccccccccccc|c}
\toprule[2pt]
\multirow{2}{*}{Method} & \multirow{2}{*}{Input} & \multirow{2}{*}{\begin{tabular}[c]{@{}c@{}}Ref.\vspace{-0.7em}\\view\end{tabular}} & \multirow{2}{*}{\begin{tabular}[c]{@{}c@{}}Recon.\vspace{-0.7em}\\-free\end{tabular}} & \multicolumn{13}{c|}{Object} & \multirow{2}{*}{\begin{tabular}[c]{@{}c@{}}Mean\vspace{-0.5em}\\(\%)$\uparrow$\end{tabular}} \\ \cline{5-17}
 &  &  &  & ape & benchwise & cam & can & cat & driller & duck & eggbox & glue & holepuncher & iron & lamp & phone & \\ \hline
Gen6D \cite{gen6d} & RGB & 200 & \textcolor{green}{\checkmark} & - & 77.0 & 66.1 & - & 60.7 & 67.4 & 40.5 & 95.7 & 87.2 & - & - & - & - & - \\
Gen6D$^\dag$ \cite{gen6d} & RGB & 200  & \textcolor{green}{\checkmark} & - & 62.1 & 45.6 & - & 40.9 & 48.8 & 16.2 & - & - & - & - & - & - & - \\ 
OnePose \cite{onepose} & RGB & 200  & \textcolor{red}{$\times$} & 11.8 & 92.6 & 88.1 & 77.2 & 47.9 & 74.5 & 34.2 & 71.3 & 37.5 & 54.9 & 89.2 & 87.6 & 60.6 & 63.6 \\ 
OnePose++ \cite{onepose++} & RGB & 200  & \textcolor{red}{$\times$} & 31.2 & 97.3 & 88.0 & 89.8 & 70.4 & 92.5 & 42.3 & 99.7 & 48.0 & 69.7 & \textbf{97.4} & \textbf{97.8} & 76.0 & 76.9 \\ 
LatentFusion \cite{latentfusion} & RGB-D & 16  & \textcolor{red}{$\times$} & \textbf{88.0} & 92.4 & 74.4 & 88.8 & 94.5 & 91.7 & 68.1 & 96.3 & 94.9 & 82.1 & 74.6 & 94.7 & 91.5 & 87.1 \\
FS6D \cite{fs6d} & RGB-D & 16  & \textcolor{green}{\checkmark} & 74.0 & 86.0 & \textbf{88.5} & 86.0 & \textbf{98.5} & 81.0 & \textbf{68.5} & \textbf{100.0} & 99.5 & \textbf{97.0} & 92.5 & 85.0 & \textbf{99.0} & 88.9 \\
Oryon \cite{corsetti2024open} &  RGB-D & 1  & \textcolor{green}{\checkmark} & 1.2 & 1.3 & 3.9 & 0.8 & 12.7  & 8.5 & 0.8 & 63.2 & 18.4 & 1.6 & 0.6 &2.9 & 11.7 & 9.8 \\
SinRef-6D (Ours) & RGB-D & 1  & \textcolor{green}{\checkmark} & 85.7 & \textbf{99.3} & 73.2 & 98.3 & 93.0 & \textbf{98.7} & 66.6 & 98.5 & 99.1 & 74.6 & 90.9 & 97.6 & 97.4 & 90.3 \\
\rowcolor{blue!6} SinRef-6D$^\wedge$ (Ours) & RGB-D & 1  & \textcolor{green}{\checkmark} & 86.3 & 99.1 & 74.7 & \textbf{98.5} & 94.5 & \textbf{98.7} & 68.1 & 98.7 & 99.5 & 75.5 & 92.5 & 97.0 & 97.8 & \textbf{90.8} \\
 \bottomrule[2pt]
\end{tabular}%
}
\label{label1}
\vspace{-1em}
\end{table*}

\noindent \textbf{Evaluation Metrics:}
1) Recall of the average point distance (ADD) that is less than 10\% of the object diameter (ADD-0.1d) \cite{add}. 2) Area under the curve (AUC) of ADD \cite{posecnn}; 3) BOP metric: Average Recall (AR) of the visible surface discrepancy (VSD), maximum symmetry-aware surface distance (MSSD), and maximum symmetry-aware projection distance (MSPD) metrics \cite{2024bop}. Specifically, we first perform a quantitative comparison using the ADD-0.1d and AUC of ADD metrics for each instance in the LineMod \cite{linemod} and YCB-V \cite{posecnn} datasets, respectively, aligning with manual reference view-based methods \cite{gen6d, onepose, onepose++,latentfusion, fs6d, predator, sun2021loftr}. Subsequently, we evaluate our results against CAD model-based methods \cite{megapose, chen2023zeropose, gigapose, SAM-6D} on five BOP datasets \cite{2024bop}, utilizing the BOP metric for a comprehensive comparison.

\begin{table}[t!]
\renewcommand{\arraystretch}{1.35}
    \centering
    \caption{Comparison of SinRef-6D with other manual reference view-based methods on the complete YCB-V dataset \cite{posecnn}, evaluated using the AUC of ADD metric.}
    \label{sup_label1}
    \resizebox{\linewidth}{!}{
    \begin{tabular}{l|cccc}
    \toprule[2pt]
         \multirow{2}{*}{Method} &  \multirow{2}{*}{\begin{tabular}[c]{@{}c@{}}PREDATOR\vspace{-0.3em}\\\cite{predator}\end{tabular}}  &  \multirow{2}{*}{\begin{tabular}[c]{@{}c@{}}LoFTR\vspace{-0.3em}\\\cite{sun2021loftr}\end{tabular}}  &  \multirow{2}{*}{\begin{tabular}[c]{@{}c@{}}FS6D-DPM\vspace{-0.3em}\\\cite{fs6d}\end{tabular}}   & \multirow{2}{*}{\begin{tabular}[c]{@{}c@{}}Ours\end{tabular}} \\
    \\
    \hline
         Reference view & 16 & 16 & 16 & 1 \\
    \hline
         002\_master\_chef\_can & 17.4 & \textbf{50.6} & 36.8 & 44.3 \\
         003\_cracker\_box & 8.3 & 25.5 & 24.5 & \textbf{34.4} \\
         004\_sugar\_box & 15.3 & 13.4 & 43.9 & \textbf{83.9} \\
         005\_tomato\_soup\_can & 44.4 & 52.9 & \textbf{54.2} & 53.7 \\
         006\_mustard\_bottle & 5.0 & 59.0 & 71.1 & \textbf{79.9} \\
         007\_tuna\_fish\_can & 34.2 & \textbf{55.7} & 53.9 & 53.8 \\
         008\_pudding\_box & 24.2 & 68.1 & \textbf{79.6} & 44.3 \\
         009\_gelatin\_box & 37.5 & 45.2 & 32.1 & \textbf{94.6} \\
         010\_potted\_meat\_can & 20.9 & 45.1 & \textbf{54.9} & 25.5 \\
         011\_banana & 9.9 & 1.6 & \textbf{69.1} & 65.0 \\
         019\_pitcher\_base & 18.1 & 22.3 & 40.4 & \textbf{88.2} \\
         021\_bleach\_cleanser & 48.1 & 16.7 & 44.1 & \textbf{72.9} \\
         024\_bowl & 17.4 & 1.4 & 0.9 & \textbf{31.7} \\
         025\_mug & 29.5 & 23.6 & 39.2 & \textbf{77.7} \\
         035\_power\_drill & 12.3 & 1.3 & 19.8 & \textbf{53.7} \\
         036\_wood\_block & 10.0 & 1.4 & \textbf{27.9} & 0.7 \\
         037\_scissors & 25.0 & 14.6 & 27.7 & \textbf{51.2} \\
         040\_large\_marker & 38.9 & 8.4 & 74.2 & \textbf{76.2} \\
         051\_large\_clamp & 34.4 & 11.2 & \textbf{34.7} & 21.4 \\
         052\_extra\_large\_clamp & \textbf{24.1} & 1.8 & 10.1 & 0.4 \\
         061\_foam\_brick & 35.5 & 31.4 & 45.8 & \textbf{56.3} \\
    \hline
         MEAN & 24.3 & 26.2 & 42.1 & \textbf{52.8} \\
    \bottomrule[2pt]
    \end{tabular}
    }
    \vspace{-0.5em}
    \label{sup_label1}
\end{table}

\subsection{Implementation Details}\label{Implementation Details}
The initial resolution of the input RGB images is $640 \times 480$, which are resized to $224 \times 224$ after detection and segmentation. Both the reference and query point clouds contain 2048 points (${N_r}$ and ${N_q}$). The point-wise feature dimension $C$ is set to 256. We use the Adam optimizer for model training with a batch size of 6, over a total of 2.4 million batches. The learning rate is adjusted using the WarmupCosineLR scheduler, starting from 0 and rapidly increasing to 0.001 during the first 1000 batches, then gradually decreasing until the end of training. The training takes $\sim$1 week on our workstation. All experiments are conducted on a single GeForce RTX 4090 GPU with an Intel Xeon Gold 6138 CPU.

\subsection{Quantitative Comparisons with SOTA Methods}\label{Quantitative Comparisons with SOTA Methods}
\subsubsection{Comparison with Manual Reference View-based Methods} Table~\ref{label1} presents a detailed performance comparison of SinRef-6D with other manual reference view-based methods \cite{gen6d, onepose, onepose++, latentfusion, fs6d, corsetti2024open} on the LineMod dataset \cite{linemod} using the ADD-0.1d metric. \textcolor{black}{For a fair comparison with Oryon \cite{corsetti2024open}, we adopt the results reported in One2Any \cite{one2any}, which directly evaluate Oryon using its pretrained model under the original setting including language input.}

\begin{table}[t!]
\renewcommand{\arraystretch}{1.3}
    \centering
    \caption{Comparison with FS6D \cite{fs6d} and FoundationPose \cite{foundationpose} under single-reference setting on LineMod and YCB-V datasets, evaluated on ADD-0.1d and AUC of ADD metrics, respectively.}
    \small
    \resizebox{\linewidth}{!}{
    \begin{tabular}{l|c|cc}
    \toprule[2pt]
         Method & Ref. view & LineMod \cite{linemod} & YCB-V \cite{posecnn} \\ \hline
         FS6D \cite{fs6d} & 1 & 77.5 & 34.7 \\
         FoundationPose \cite{foundationpose} & 1 & 87.9 & 47.5 \\
         SinRef-6D (Ours) & 1 & \textbf{90.3} & \textbf{52.8} \\ 
    \bottomrule[2pt]
    \end{tabular}
    }
    \vspace{-0.5em}
    \label{addlabel1}
\end{table}

\par We evaluate SinRef-6D using two reference view selection strategies, as described in Sec. \ref{Initialization}. \emph{In the first setting}, we follow the same rendering protocol as GigaPose \cite{gigapose}, where a single reference view is randomly sampled from the \emph{50th} to the \emph{120th} viewpoint in its rendering sequence. Under this setting, SinRef-6D achieves a mean accuracy of 90.3\%, outperforming OnePose \cite{onepose}, OnePose++ \cite{onepose++}, and Oryon \cite{corsetti2024open}, and achieving comparable accuracy to LatentFusion \cite{latentfusion} and FS6D \cite{fs6d}. It is worth noting that all the comparison methods rely on dense reference views. Specifically, Gen6D \cite{gen6d}, OnePose \cite{onepose}, and OnePose++ \cite{onepose++} require 200 reference views, while LatentFusion \cite{latentfusion} and FS6D \cite{fs6d}, which utilize RGB-D inputs like SinRef-6D, still need 16 reference views. Moreover, Gen6D \cite{gen6d} depends on a template matching process, while OnePose \cite{onepose}, OnePose++ \cite{onepose++}, and LatentFusion \cite{latentfusion} require reconstructing the 3D object representation prior to pose estimation. These additional requirements will increase model complexity and reduce overall efficiency. \emph{In the second setting}, we manually select an occlusion-free reference view from the training/testing sets that closely approximates the robotic manipulation viewpoint. This strategy leads to a slightly higher accuracy of 90.8\%, which we attribute to the improved viewpoint alignment between the reference and query images. The experimental results of these two reference view selection settings collectively demonstrate that SinRef-6D does \emph{not rely on a carefully selected} reference view and \emph{remains robust to variations} in the selection of the reference view.

\begin{table*}[!t]
\renewcommand{\arraystretch}{1.2}
\centering
\caption{Comparison of SinRef-6D with CAD model-based methods on the LM-O \cite{linemod-o}, TUD-L \cite{2024bop}, IC-BIN \cite{IC-BIN}, HB \cite{HB}, and YCB-V \cite{posecnn} datasets. We leverage the BOP metric and the mean time across all datasets for evaluation. $^\wedge$ denotes that the reference view is manually selected from the corresponding dataset to approximate the robotic manipulation viewpoint during training and testing. $^{*}$ denotes using the pose refinement method of MegaPose \cite{megapose}. $^{\dagger}$ means that directly test SAM-6D \cite{SAM-6D} uses a single reference view. $^{\ddag}$ means that retrain and then test SAM-6D \cite{SAM-6D} uses a single reference view, with other settings unchanged.}
\fontsize{6}{8.5}\selectfont
\resizebox{\linewidth}{!}{
\begin{tabular}{l|cc|c|ccccc|c|c}
\toprule[2pt]
\multirow{2}{*}{Method} & \multirow{2}{*}{Input} & \multirow{2}{*}{\begin{tabular}[c]{@{}c@{}}CAD model\vspace{-0.5em}\\-free\end{tabular}} & \multirow{2}{*}{Detection / Segmentation} & \multicolumn{5}{c|}{Dataset} & \multirow{2}{*}{\begin{tabular}[c]{@{}c@{}}Mean\vspace{-0.5em}\\(\%)$\uparrow$\end{tabular}} & \multirow{2}{*}{\begin{tabular}[c]{@{}c@{}}Time\vspace{-0.5em}\\(s)$\downarrow$\end{tabular}} \\ \cline{5-9}
 &  &  &  & LM-O & TUD-L & IC-BIN & HB & YCB-V &  \\ \hline
SinRef-6D (Ours) & RGB-D & \textcolor{green}{\checkmark} & \multirow{2}{*}{Single Ref.-based CNOS} & 48.4 & 62.5 & 31.6 & 50.7 & 56.9 & 50.0 & 0.7 \\
SinRef-6D$^\wedge$ (Ours) & RGB-D & \textcolor{green}{\checkmark} &  & 51.2 & 65.3 & 32.7 & 52.9 & 58.4 & 52.1 & 0.7 \\ \hline
MegaPose \cite{megapose} & RGB & \textcolor{red}{$\times$} & \multirow{9}{*}{Mask R-CNN \cite{maskrcnn}} & 18.7 & 20.5 & 15.3 & 18.6 & 13.9 & 17.4 & 25.6 \\
MegaPose* \cite{megapose} & RGB & \textcolor{red}{$\times$} &  & 53.7 & 58.4 & 43.6 & 72.9 & 60.4 & 57.8 & - \\
MegaPose* \cite{megapose} & RGB-D & \textcolor{red}{$\times$} &  & 58.3 & 71.2 & 37.1 & \textbf{75.7} & 63.3 & 61.1 & 93.3 \\
ZeroPose \cite{chen2023zeropose} & RGB-D & \textcolor{red}{$\times$} &  & 26.1 & 61.1 & 24.7 & 38.2 & 29.5 & 35.9 & - \\
ZeroPose* \cite{chen2023zeropose} & RGB-D & \textcolor{red}{$\times$} &  & 56.2 & 87.2 & 41.8 & 68.2 & 58.4 & 62.4 & - \\
SAM-6D$^{\dagger}$ \cite{SAM-6D} & RGB-D & \textcolor{red}{$\times$} &  & 12.9 & 37.9 & 11.2 & 25.2 & 22.4 & 21.9 & \textbf{0.3} \\
SAM-6D$^{\ddag}$ \cite{SAM-6D} & RGB-D & \textcolor{red}{$\times$} &  & 53.7 & 38.4 & 26.3 & 53.2 & 60.1 & 46.3 & \textbf{0.3} \\
SinRef-6D (Ours) & RGB-D & \textcolor{green}{\checkmark} &  & 61.8 & 88.9 & 44.0 & 63.3 & 65.1 & 64.6 & 0.4 \\ 
\rowcolor{blue!6} SinRef-6D$^\wedge$ (Ours) & RGB-D & \textcolor{green}{\checkmark} &  & \textbf{62.0} & \textbf{90.4} & \textbf{44.5} & 64.0 & \textbf{65.9} & \textbf{65.4} &  0.4\\
\hline
MegaPose \cite{megapose} & RGB & \textcolor{red}{$\times$} & \multirow{9}{*}{CNOS (FastSAM) \cite{cnos}} & 22.9 & 25.8 & 15.2 & 25.1 & 28.1 & 23.4 & 16.6\\
MegaPose* \cite{megapose} & RGB & \textcolor{red}{$\times$} &  & 49.9 & 65.3 & 36.7 & 65.4 & 60.1 & 55.5 &  33.9\\
ZeroPose* \cite{chen2023zeropose} & RGB-D & \textcolor{red}{$\times$} &  & 53.8 & \textbf{83.5} & 39.2 & 65.3 & 65.3 & 61.4 &  17.6\\
GigaPose \cite{gigapose} & RGB & \textcolor{red}{$\times$} &  & 29.6 & 30.0 & 22.3 & 34.1 & 27.8 & 28.8 &  \textbf{0.4}\\
GigaPose* \cite{gigapose} & RGB & \textcolor{red}{$\times$} &  & \textbf{59.8} & 63.1 & \textbf{47.3} & \textbf{72.2} & \textbf{66.1} & \textbf{61.7} &  8.5\\
SAM-6D$^{\dagger}$ \cite{SAM-6D} & RGB-D & \textcolor{red}{$\times$} &  & 10.4 & 30.1 & 9.4 & 29.0 & 21.8 & 20.1 & 1.2 \\
SAM-6D$^{\ddag}$ \cite{SAM-6D} & RGB-D & \textcolor{red}{$\times$} &  & 53.9 & 32.2 & 25.0 & 55.4 & 59.1 &  45.1 & 1.1 \\
SinRef-6D (Ours) & RGB-D & \textcolor{green}{\checkmark} &  & 56.5 & 77.4 & 35.9 & 61.0 & 62.2 & 58.6 &  1.5\\
\rowcolor{blue!6} SinRef-6D$^\wedge$ (Ours) & RGB-D & \textcolor{green}{\checkmark} &  & 56.2 & 78.6 & 37.1 & 62.1 & 62.9 & 59.4 &  1.5\\
 \bottomrule[2pt]
\end{tabular}%
}
\vspace{-0.25em}
\label{label2}
\end{table*}

\par Additionally, we evaluate SinRef-6D on the complete YCB-V \cite{posecnn} dataset using the AUC of ADD metric, with per-object accuracy results summarized in Tab.~\ref{sup_label1}. These quantitative results further validate the advantages of SinRef-6D. Also, we note that objects with flat or geometrically complex structures, such as the extra large clamp, tend to yield lower accuracy, as they are difficult to perceive accurately from a single reference view. Furthermore, we set the number of reference views to 1 for both FS6D \cite{fs6d} and FoundationPose \cite{foundationpose} and experiment with the LineMod and YCB-V datasets, \emph{using their pre-trained models and keeping other settings unchanged}. In particular, for FoundationPose, we adopt its ablation setting by reconstructing the unseen object 3D model from a single reference, while keeping the downstream hypothesis generation and pose selection components fixed. Notably, FoundationPose utilizes a higher-quality synthetic dataset generated with diffusion model and requires time-consuming 3D reconstruction (making it $ \sim $10 times slower than SinRef-6D). The results, as shown in Tab.~\ref{addlabel1}, further reinforce the advantages of SinRef-6D in the single-reference setting.

\par Overall, the above experimental results demonstrate the effectiveness of the proposed SinRef-6D task setup and framework, achieving competitive 6-DoF pose estimation accuracy for unseen objects using only a single reference view, without relying on the reconstruction of a 3D object representation.

\subsubsection{Comparison with CAD Model-based Methods} Table~\ref{label2} presents performance comparisons of SinRef-6D with CAD model-based methods \cite{megapose, chen2023zeropose, gigapose, SAM-6D} on five popular datasets using the BOP metric (AR metric of VSD, MSSD, and MSPD). We also conduct experiments under two reference view selection strategies, as outlined in Sec. \ref{Initialization}. The first strategy adopts the same rendering manner as GigaPose \cite{gigapose},
and the second strategy involves manually choosing an occlusion-free view from the training/testing set that closely aligns with the robotic manipulation view. In addition, we further evaluate SinRef-6D under three different detection/segmentation methods to assess its robustness across perception inputs. Specifically, we first show the experimental results using only our single reference view as the template image for CNOS segmentation \cite{cnos} in the first two rows.

\par When Mask R-CNN \cite{maskrcnn} is used to segment the input images, SinRef-6D achieves 64.6\% AR across five evaluation datasets, outperforming MegaPose \cite{megapose} and ZeroPose \cite{chen2023zeropose}. Remarkably, this accuracy even exceeds the performance of these two methods after they leverage the refinement method introduced by MegaPose \cite{megapose}.
Moreover, we evaluate SAM-6D \cite{SAM-6D} also with a single reference view under two training settings while keeping other settings (e.g., coarse-to-fine refine) unchanged.
Next, when we employ zero-shot CNOS \cite{cnos} for segmentation, SinRef-6D also achieves a competitive AR of 58.6\% across the five datasets.
We also note an increase in inference time. This is primarily because CNOS often segments multiple instances for the same object, resulting in repeated pose estimations.
Note that the comparison methods all rely on textured CAD models, requiring specialized equipment for acquisition. Additionally, the refinement process of MegaPose \cite{megapose} is slow and also relies on object CAD models (that is why we do not use it for refinement). In contrast, SinRef-6D is CAD model- and refinement-free, offering enhanced scalability and efficiency. In general, SinRef-6D demonstrates performance on par with CAD model-based methods, while operating in a CAD model-free setup, showcasing its effectiveness and scalability.

\begin{table}[!t]
\renewcommand{\arraystretch}{1.3}
\centering
\caption{\textcolor{black}{Comparison with other single-reference methods.}}
\fontsize{8}{10}\selectfont
\begin{tabular}{l|c|cc}
\toprule[2pt]
\multirow{2}{*}{Method} & \multicolumn{1}{c|}{LM} & LM-O & TUD-L  \\ \cline{2-4}
 &  ADD-0.1d & \multicolumn{2}{c}{AR} \\ \hline
One2Any \cite{one2any} & 52.6 & - & - \\
Any6D \cite{any6d} & - & 28.6 & - \\
UNOPose \cite{unopose} & - & 56.0 & 67.1 \\
Ours  & \textbf{90.3} & \textbf{56.5} & \textbf{77.4} \\
 \bottomrule[2pt]
\end{tabular}%
\label{R1addlabel1}
\end{table}

\par Similarly, across all three detection/segmentation methods, we observe that manually selected reference views consistently yield slightly higher accuracy than randomly rendered ones. We argue that this is due to their closer alignment with the robotic manipulation viewpoint and reduced ambiguity. Overall, these results not only highlight the robustness of SinRef-6D to different detection/segmentation pipelines, but also further confirm its resilience to variations in reference view selection.

\subsubsection{Comparison with Single-Reference Methods}
Table \ref{R1addlabel1} shows explicit quantitative comparisons with the three most closely related single-reference methods discussed in Sec. \ref{Relative Pose Estimation Methods} (One2Any \cite{one2any}, Any6D \cite{any6d}, and UNOPose \cite{unopose}). To ensure a fair comparison, all methods use the same segmentation and evaluation protocols: ground-truth segmentation on the LM dataset, and CNOS segmentation \cite{cnos} on the LM-O and TUD-L datasets; ADD-0.1d is used for LM, while AR is used for LM-O and TUD-L, consistent with prior work. The results of One2Any and Any6D are taken directly from their original papers, while UNOPose is evaluated using its pretrained model under CNOS segmentation (same as SinRef-6D). These results show that SinRef-6D achieves more accurate pose estimation performance.

\subsection{Qualitative Analysis}\label{Qualitative Analysis}
\subsubsection{Comparison with Manual Reference View-based Methods} 
\begin{figure}[t!]
    \centering
    \includegraphics[width=\linewidth]{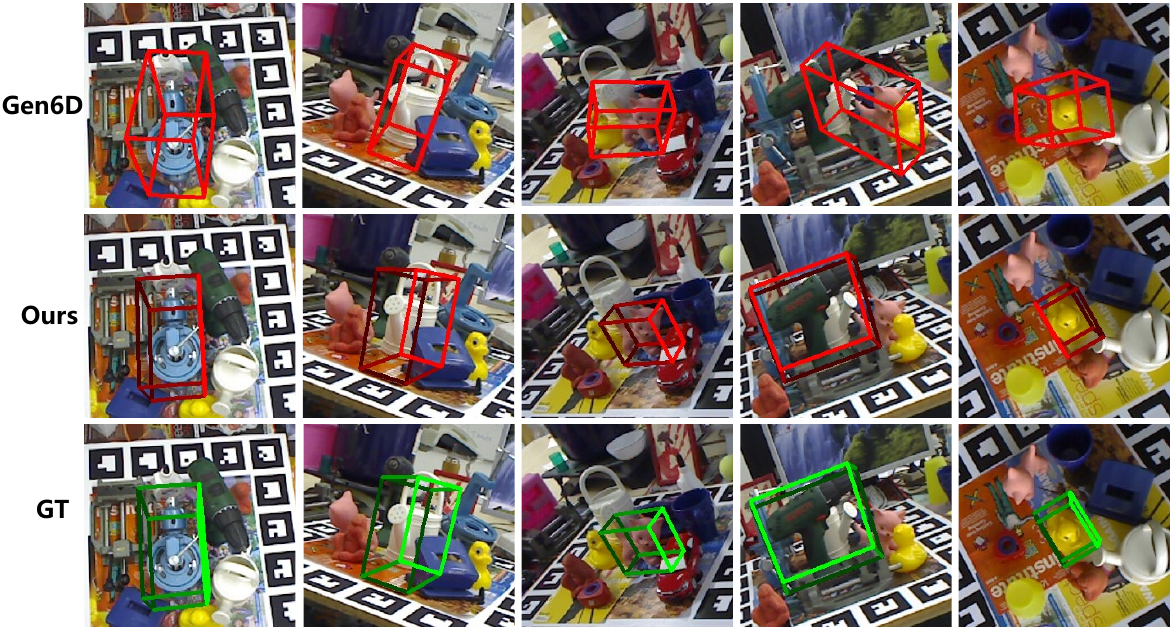}
    \vspace{-1em}
    \caption{The qualitative comparison results on the LineMod dataset \cite{linemod} are presented, visualizing the outputs of Gen6D \cite{gen6d}, our SinRef-6D, and ground truth from top to bottom.
    }
    \label{Fig3}
    \vspace{-0.5em}
\end{figure}

The comparison between SinRef-6D and Gen6D \cite{gen6d} on the LineMod dataset \cite{linemod} is presented in Fig.~\ref{Fig3}. These experimental results highlight the superior performance of our method, which can perform unseen object pose estimation in cluttered scenes. Specifically, Gen6D \cite{gen6d} relies on dense reference views and template matching to estimate object poses, which requires full coverage of the reference view angles. When the number of reference views is limited or of low quality, pose estimation errors will significantly increase. In contrast, SinRef-6D abandons template matching and instead leverages iterative alignment between the reference and query views, enabling effective pose estimation of unseen objects using only a single reference view.

\begin{figure*}[t!]
    \centering
    \includegraphics[width=\linewidth]{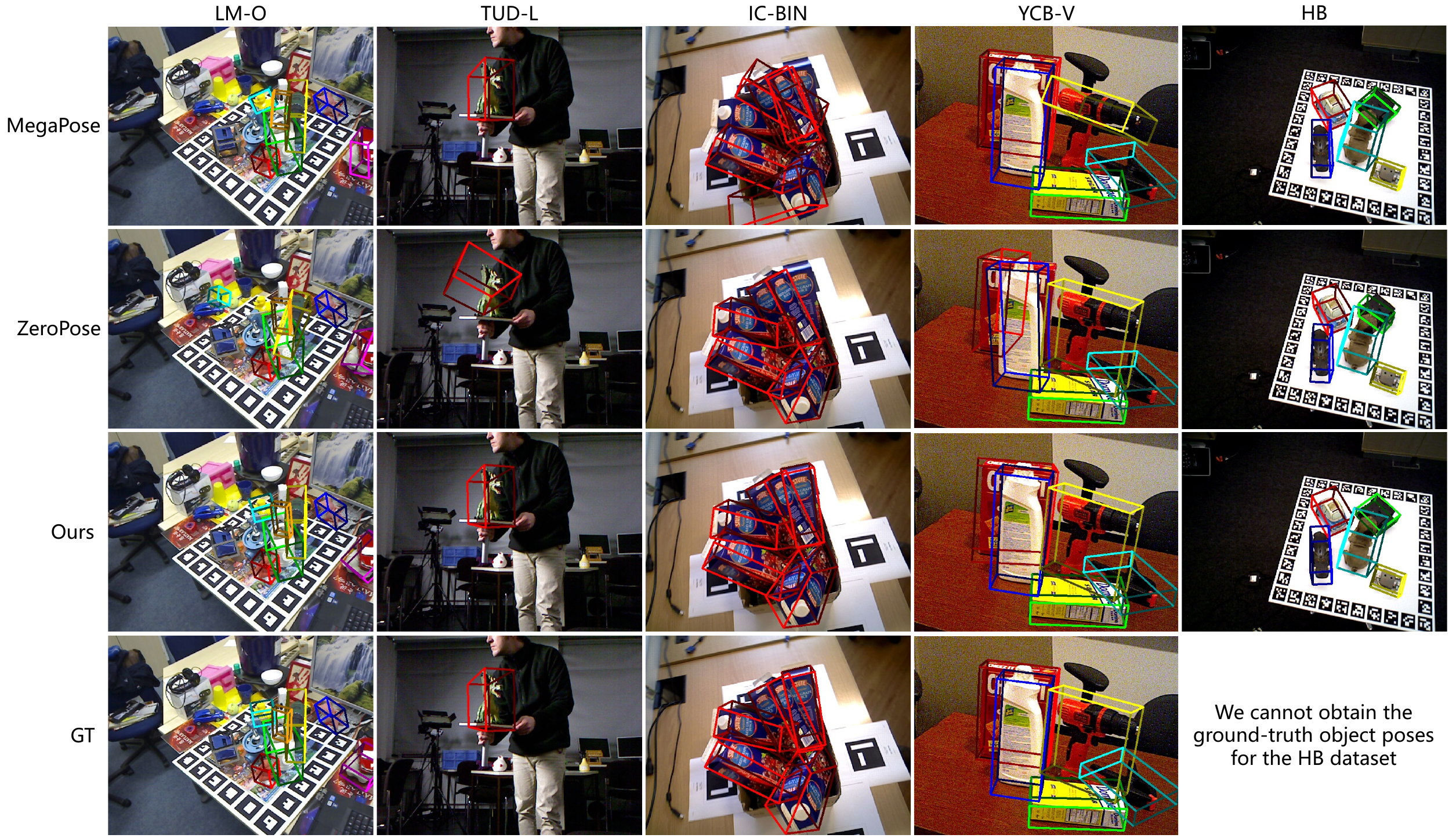}
    \vspace{-1em}
    \caption{The qualitative comparison results on the LM-O \cite{linemod-o}, TUD-L \cite{2024bop}, IC-BIN \cite{IC-BIN}, and YCB-V \cite{posecnn} datasets. We visualize the results of MegaPose \cite{megapose}, ZeroPose \cite{chen2023zeropose}, our SinRef-6D, and ground truth from top to bottom. Note that we cannot obtain the ground-truth poses for the HB dataset \cite{HB}, as its evaluation is conducted on the official BOP Challenge \cite{2024bop}. Additional qualitative results are presented at our project \href{https://paperreview99.github.io/SinRef-6DoF-Robotic}{homepage}.
    }
    \label{Fig4}
    \vspace{-0.5em}
\end{figure*}

\subsubsection{Comparison with CAD Model-based Methods} Figure~\ref{Fig4} compares SinRef-6D with MegaPose \cite{megapose} and ZeroPose \cite{chen2023zeropose} across five evaluation datasets, all of which use RGB-D input and CNOS \cite{cnos} for segmentation. These datasets cover a diverse range of scenes and unseen objects. The experimental results further demonstrate that our method outperforms the comparison methods in terms of robustness, effectively estimating the 6-DoF pose of unseen objects even in challenging scenes with occlusions and clutter. Notably, both MegaPose \cite{megapose} and ZeroPose \cite{chen2023zeropose} require textured CAD models of these unseen objects, with MegaPose \cite{megapose} also depending on the time-consuming render-and-compare process. In contrast, SinRef-6D is a simple-yet-effective CAD-free method, offering greater scalability.

\begin{figure}[t!]
    \centering
    \includegraphics[width=\linewidth]{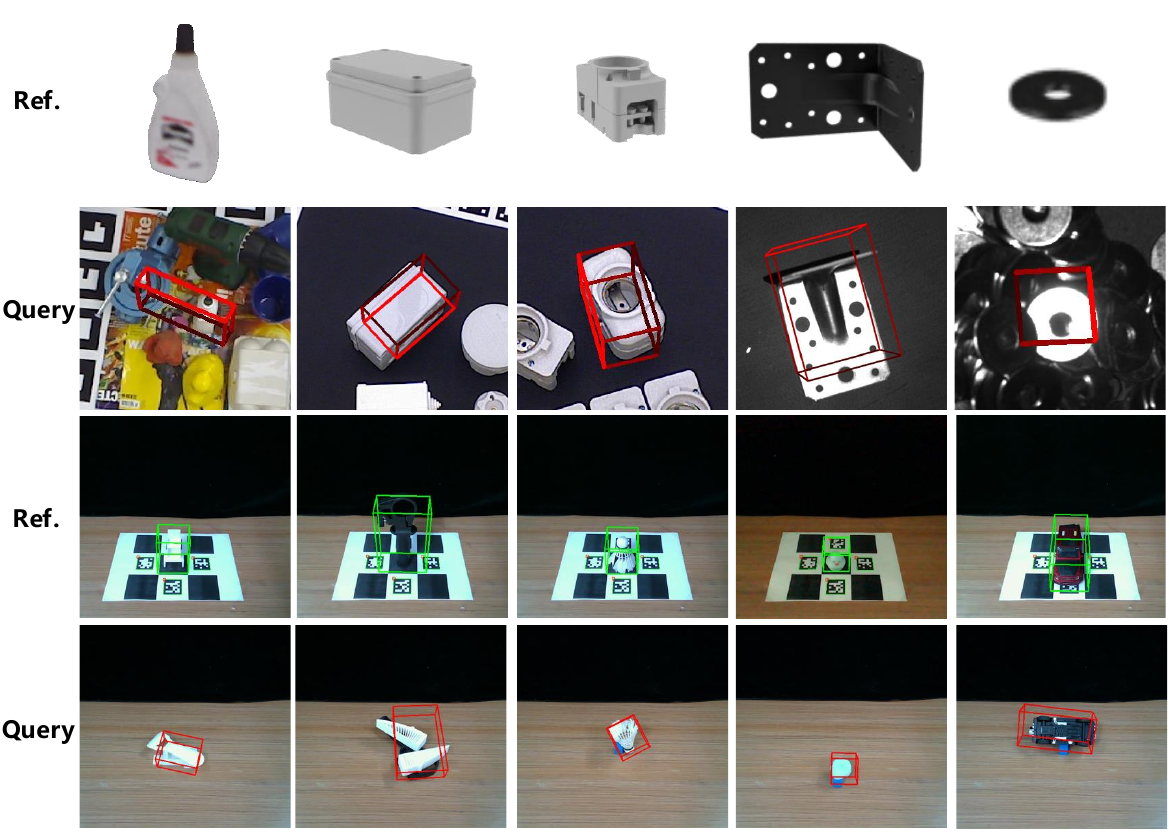}
    \vspace{-1.5em}
    \caption{\textcolor{black}{Failure cases on public datasets (top two rows) and real world (bottom two rows). The top two rows show single reference views (randomly selected RGB-D images as described in Sec.~\ref{Initialization}) and the estimated object pose in query views. As observed, the accuracy of SinRef-6D decreases when the query view is a top-down view or the object is a reflective metal. The bottom two rows illustrate the labeled reference views and representative failure cases arising from particularly challenging scenarios, such as severe occlusion/self-occlusion and large viewpoint gaps between the reference and query views.}
    }
    \label{Fig5}
    \vspace{-0.5em}
\end{figure}

\subsubsection{Failure Cases Analysis} 
Since SinRef-6D only uses a single reference view captured from an oblique angle, pose estimation accuracy may decrease when the query view does not adequately capture the object's geometric features, such as in top-down views. Furthermore, for objects with incomplete depth information, like reflective metals or transparent materials, establishing accurate point-wise alignments becomes challenging, leading to a decrease in pose estimation accuracy. The visualization of some failure cases is shown in Fig.~\ref{Fig5}. As a result, we do not evaluate the T-LESS \cite{tless} (includes many top-down views) and ITODD \cite{itodd} (features top-down views and metallic objects) datasets. \textcolor{black}{Moreover, we provide an explicit analysis of real-world failure cases and observe a performance degradation under non-planar object placements, large reference-query viewpoint discrepancies, and severe occlusions, as illustrated in the last two rows of Fig.~\ref{Fig5}. Our future work will focus on enhancing the robustness of SinRef-6D in such challenging scenes and objects.}

\begin{figure*}[t!]
    \centering
    \includegraphics[width=\linewidth]{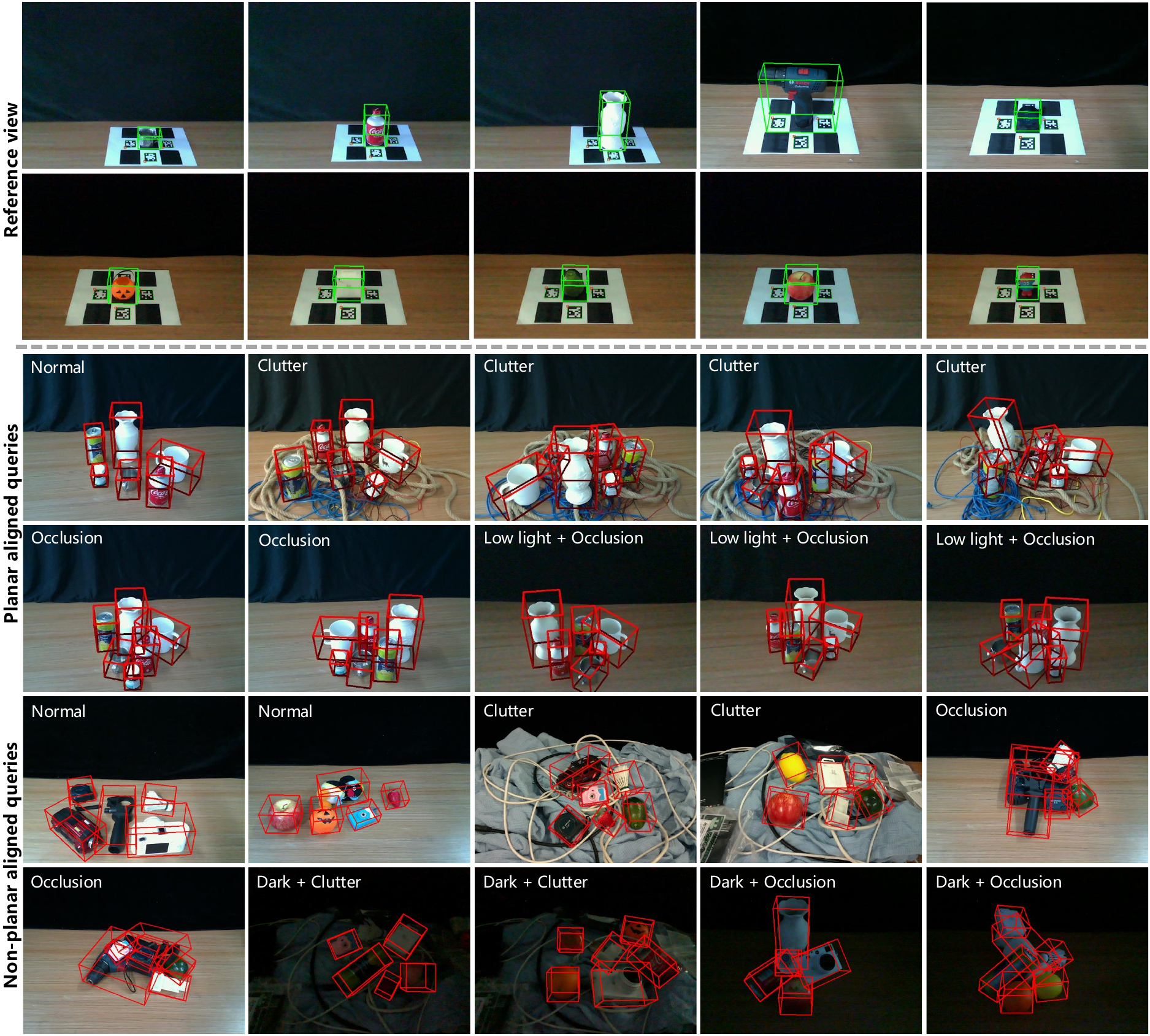}
    \vspace{-1em}
    \caption{\textcolor{black}{Qualitative results on real-world robotic grasping scenarios. \emph{Top two rows}: Single reference view capture and annotation from the robotic manipulation viewpoint for several unseen objects. \emph{Middle two rows}: Unseen object 6-DoF pose estimation in planar aligned query views. \emph{Bottom two rows}: Unseen object 6-DoF pose estimation in non-planar aligned query views. These include some challenging scenes commonly encountered in robotic grasping, including clutter, occlusion, low light, and dark conditions.}}
    \label{graspsence}
    \vspace{-0.5em}
\end{figure*}

\subsection{Real-world Robotic Grasping}\label{Real-world Robotic Grasping}
\subsubsection{Qualitative validation on real-world robotic grasping scenarios}
To evaluate the effectiveness of SinRef-6D in estimating the 6-DoF poses of unseen objects in real-world robotic grasping scenarios, we mount an Intel RealSense L515 RGB-D camera on a robotic arm and conduct experiments across four representative scenes: normal, clutter, occlusion, and low light. \textcolor{black}{Qualitative results are shown in Fig. \ref{graspsence}. Specifically, the robot first captures a reference image of an unseen object from its manipulation viewpoint, which is then annotated using our custom-developed semi-automatic annotator within one minute (top two rows). This reference is used as prior knowledge for zero-shot unseen object 6-DoF pose estimation from other viewpoints. We place multiple unseen objects in some challenging scenes and evaluate the robustness of SinRef-6D under realistic grasping conditions (middle two rows). Since the robustness to non-planar object placements is critical for real-world robotic deployments, we substantially introduce some objects placed in inverted or non-planar configurations (bottom two rows). In general, these experiments validate the effectiveness of SinRef-6D and demonstrate its potential for downstream robotic grasping tasks.}

\subsubsection{Real-world robotic grasping}
\begin{figure*}[t!]
    \centering
    \includegraphics[width=\linewidth]{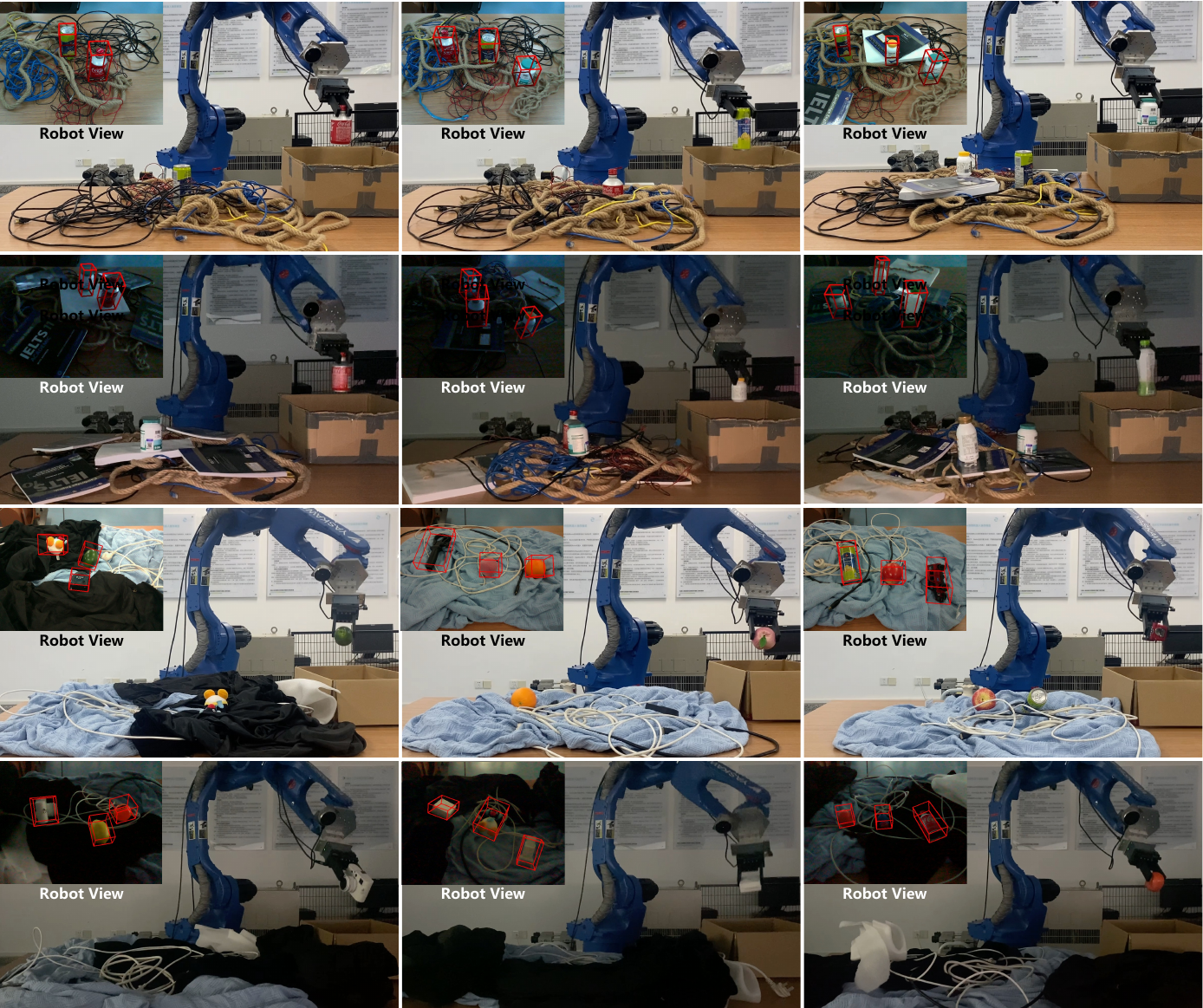}
    \vspace{-1em}
    \caption{\textcolor{black}{Real-world robotic grasping visualizations. The top two rows and the bottom two rows represent planar aligned and non-planar aligned object grasping, respectively. These scenes contain clutter and low light. The top-left corner in each image shows the estimated 6-DoF poses of the unseen objects from the robotic manipulation viewpoint. The complete grasping video can be seen through our project \href{https://paperreview99.github.io/SinRef-6DoF-Robotic}{homepage}.}}
    \label{grasp}
\end{figure*}

To evaluate the applicability of SinRef-6D to unseen object 6-DoF robotic grasping and to validate the effectiveness of the developed hardware-software robotic system (described in Sec. \ref{Developed Robotic Grasping System}), we integrate SinRef-6D on the system to perform real-world grasping experiments. Due to the mechanical limitations of the gripper, we select some graspable objects placed in randomly cluttered scenes for testing. Some representative qualitative results are shown in Fig. \ref{grasp}. Specifically, we first estimate the 6-DoF poses of these unseen objects (top-left) under both normal and low light scenarios, and then execute sequential robotic grasping based on the estimated poses. \textcolor{black}{To further assess robustness under more challenging conditions, we additionally include grasping demonstrations involving geometrically irregular unseen objects placed in non-planar configurations and under large reference-query viewpoint discrepancies, as shown in the bottom two rows of Fig. \ref{grasp}. In these experiments, object poses are randomly placed, while scene clutter is generated by randomly moving non-target, non-rigid items such as blankets and cables. Quantitatively, we conduct a total of 200 real-world grasping trials, evenly divided between planar placements and challenging non-planar object configurations (100 trials each). Each scene contains two or three randomly placed unseen objects, achieving overall success rates of 85\% and 74\%, respectively, where a trial is considered successful only if all objects in the scene are successfully grasped.} In summary, these real-world experiments not only demonstrate the effectiveness of the proposed task formulation and method for unseen object 6-DoF robotic grasping, but also validate the robustness and practicality of our integrated robotic system.

\subsection{Ablation Study}\label{Ablation Study}
We investigate the effectiveness of several main components within SinRef-6D, as well as the impact of the number of iterations for point-wise alignment during both training and inference. Additionally, we further conduct multiple experiments with randomly selected reference views to demonstrate the robustness of our method to such variations.

\subsubsection{Effectiveness of Main Components}
We conduct a thorough ablation study on the main components in SinRef-6D to verify their effectiveness. Specifically, we first remove the RGB image component from SinRef-6D, meaning that RGB images are not used during either training or inference. The experimental results (row A of Tab. \ref{label3}) show that RGB images play a crucial role in enhancing the accuracy of point-wise alignment. Next, we eliminate the point cloud focalization process, directly feeding the reference and query point clouds ${P_r}$ and ${P_q}$ from part (B) of Fig. \ref{framework} into the model. Therefore, the iterative training of the GeoTransformer is also removed. The experimental results (row B of Tab. \ref{label3}) reveal a substantial drop in performance, underscoring the significance of the focalization step. We attribute this decline to the larger numerical discrepancies that may arise between the reference and query point clouds without the focalization process. \textcolor{black}{In addition, we only train a single GeoTransformer for point cloud iterative alignment (row C of Tab. \ref{label3}). Although the parameter count is slightly reduced, the performance degrades significantly. This is because a single alignment model struggles to simultaneously specialize in handling large initial pose discrepancies and accurately refining small residual errors.} Further, we replace the proposed RGB SSM with DINOv2 \cite{dinov2} (a large vision foundation model) and Vision Transformer \cite{vit}, and also replace Point SSM with Point Transformer \cite{pointtransformer}. The corresponding results (rows D, E, and F of Tab. \ref{label3}) confirm the effectiveness of SSMs in improving the accuracy of point-wise alignment while reducing the model parameter count.

\subsubsection{Impact of the Number of Iterations for Point-wise Alignment}
To achieve precise object pose estimation, it is essential to perform point-wise iterative alignment of the reference and query point clouds. Identifying an optimal balance between accuracy and computational efficiency requires careful selection of iteration counts during training and inference. \textcolor{black}{The training iterations refer to the number of times the GeoTransformer weights are updated during training (\emph{i.e.}, $K$ in Eq.~(\ref{equation8}))}, while the inference iterations indicate the number of GeoTransformer iterations during inference. If the inference iterations exceed the training iterations, the difference represents how many times the last GeoTransformer weights from training are repeatedly applied. The results, summarized in Tab. \ref{label4}, reveal a trade-off among accuracy, speed, and model complexity. \textcolor{black}{As shown in the first row, using a single GeoTransformer for all alignment iterations leads to degraded performance in later refinement stages. We attribute this to a distribution mismatch between the training data, which is dominated by large pose discrepancies, and the inputs encountered during refinement, where the alignment error is much smaller. This observation motivates the use of a separate refinement model specialized for local alignment. Alternative designs, such as conditioning a single alignment model on the iteration index or estimated pose discrepancy, or adopting mixture-of-experts or curriculum learning strategies to handle different alignment regimes, could potentially address this dynamic range challenge and be explored in future work.} Based on these ablations, we select 2 training and 3 inference iterations as the optimal configuration to achieve the best balance.

\begin{table}[t!]
\renewcommand{\arraystretch}{1.3}
    \centering
    \caption{Ablation of SinRef-6D components on the YCB-V dataset \cite{posecnn}. Param. means total model parameters.}
    \fontsize{8}{10}\selectfont
    \begin{tabular}{l|l|c|c}
    \toprule[2pt]
       Row & Method & AR $\uparrow$ & Param. (M) $\downarrow$\\
    \hline
       A & w/o RGB & 39.5 & 138.8 \\
       B & w/o Points Focalization & 0.0 & 643.6 \\
       \textcolor{black}{C} & \textcolor{black}{only one GeoTransformer} & \textcolor{black}{36.5} & \textcolor{black}{643.6}\\
       D & RGB SSM $\to$ DINOv2 \cite{dinov2} & 56.9 & 1238.8 \\ 
       E & RGB SSM $\to$ ViT \cite{vit} & 52.8 & 976.7 \\ 
       F & Point SSM $\to$ PT \cite{pointtransformer} & 60.9 & 708.6 \\
       G & Full Model & 62.2 & 691.8 \\
    \bottomrule[2pt]
    \end{tabular}
    \vspace{-0.5em}
    \label{label3}
\end{table}

\begin{table}[t!]
\renewcommand{\arraystretch}{1.3}
    \centering
    \caption{Ablation on the number of iterations for point-wise alignment. Note that the time is only on the YCB-V dataset \cite{posecnn}.}
    \fontsize{8}{10}\selectfont
    \begin{tabular}{p{25mm}<{\centering} | p{8mm}<{\centering} p{8mm}<{\centering} p{8mm}<{\centering} p{8mm}<{\centering}}
    \toprule[2pt]
         \multirow{2}{*}{Training iterations} & \multicolumn{4}{c}{Inference iterations, AR (\%) $\uparrow$} \\ \cline{2-5}
         & 1 & 2 & 3 & 4 \\
    \hline
         1 & 37.8  & 35.0  & 36.5  & 35.9 \\
         2 & - & 61.7  &  62.2 &  62.5 \\  
         3 & - & - &  62.2 &  62.6 \\ 
    \hline
         Time (s) $\downarrow$& 0.70  & 0.99  & 1.26  & 1.51  \\  
    \bottomrule[2pt]
    \end{tabular}
    \vspace{-0.25em}
    \label{label4}
\end{table}

\subsubsection{Impact of Random Reference View Selection} 
We investigate the effect of our random reference view selection method (see Sec. \ref{Initialization} for details) on the performance of the object 6-DoF pose estimation. \textcolor{black}{Unlike the reference view variations evaluated in Tabs.~\ref{label1} and \ref{label2}, which compare different reference acquisition strategies, the objective here is to assess the robustness of the proposed model under stochastic reference view selection. Specifically, we conduct 20 tests on both the YCB-V and LM-O datasets, where in each test the reference view is randomly sampled and rendered from the viewpoints \emph{50th} to \emph{120th} defined in GigaPose \cite{gigapose}, following the same reference view sampling protocol used during training.} Experimental results in Tab. \ref{label7} show that the variance of multiple experiments is small, which means that SinRef-6D is robust to random sampled reference views.

\begin{table}[t!]
\renewcommand{\arraystretch}{1.3}
    \centering
    \caption{Ablation of random reference view selection, we report the mean and variance of 20 times experiments on the BOP metric.}
    \fontsize{8}{10}\selectfont
    \begin{tabular}{l|c|c|c}
    \toprule[2pt]
       Dataset & Times & Mean & Variance\\
    \hline
       YCB-V \cite{posecnn} & 20 & 62.10 & 0.31 \\
       LM-O \cite{linemod-o} & 20 & 56.58 & 0.43 \\
    \bottomrule[2pt]
    \end{tabular}
    \vspace{-0.25em}
    \label{label7}
\end{table}

\section{Conclusion}\label{Conclusion}
We proposed SinRef-6D, a simple-yet-effective task setup and framework to tackle the challenges of unseen object 6-DoF pose estimation in existing methods that rely on textured CAD models or dense reference views. SinRef-6D solely requires a single reference view and iteratively establishes point-wise alignment between the reference and query views in the object coordinate system using our proposed SSMs, eliminating the reliance on a CAD model and substantially enhancing the scalability for real-world applications.
In addition, we developed a complete hardware-software robotic grasping system tailored to the proposed task setup and framework. This system integrates hand-eye and tool calibration, a semi-automatic single reference annotator, a pre-trained SinRef-6D model, and grasping strategy and control. Extensive experiments on six benchmarks and real-world robotic grasping scenarios demonstrate the superior scalability of SinRef-6D and the effectiveness of our developed robotic system. 

\noindent \textbf{Limitation and Future Work:} While the proposed framework demonstrates robust real-world pick-and-place performance, the current grasping policy is deliberately simple and does not fully exploit the estimated 6-DoF object pose. Our future work will investigate tighter coupling between pose estimation and downstream manipulation, including pose-conditioned grasp planning and more dexterous manipulation tasks.

\bibliographystyle{IEEEtran}
\bibliography{reference}

@article{chen2025sam-3d,
  title={Sam 3d: 3dfy anything in images},
  author={Chen, Xingyu and Chu, Fu-Jen and Gleize, Pierre and Liang, Kevin J and Sax, Alexander and Tang, Hao and Wang, Weiyao and Guo, Michelle and Hardin, Thibaut and Li, Xiang and others},
  journal={arXiv preprint arXiv:2511.16624},
  year={2025}
}

@inproceedings{wang2025vggt,
  title={Vggt: Visual geometry grounded transformer},
  author={Wang, Jianyuan and Chen, Minghao and Karaev, Nikita and Vedaldi, Andrea and Rupprecht, Christian and Novotny, David},
  booktitle={Proceedings of the Computer Vision and Pattern Recognition Conference},
  pages={5294--5306},
  year={2025}
}

@article{stary2025understanding,
  title={Understanding multi-view transformers},
  author={Stary, Michal and Gaubil, Julien and Tewari, Ayush and Sitzmann, Vincent},
  journal={arXiv preprint arXiv:2510.24907},
  year={2025}
}

@article{chen2025tro,
  title={A multi-level similarity approach for single-view object grasping: matching, planning, and fine-tuning},
  author={Chen, Hao and Kiyokawa, Takuya and Hu, Zhengtao and Wan, Weiwei and Harada, Kensuke},
  journal={IEEE Transactions on Robotics},
  year={2025}
}

@article{tro2024_survey,
  title={Challenges for monocular 6-d object pose estimation in robotics},
  author={Bauer, Dominik and H{\"o}nig, Peter and Weibel, Jean-Baptiste and Garc{\'\i}a-Rodr{\'\i}guez, Jos{\'e} and Vincze, Markus and others},
  journal={IEEE Transactions on Robotics},
  volume={40},
  pages={4065-4084},
  year={2024}
}

@article{tro2025,
  title={Fast and accurate 6-d object pose refinement via implicit surface optimization},
  author={Pang, Bo and Zhai, Deming and Zhen, Jianan and Wang, Long and Liu, Xianming},
  journal={IEEE Transactions on Robotics},
  volume={41},
  pages={3129-3142},
  year={2025}
}

@article{talak_tro2023,
  title={Certifiable object pose estimation: Foundations, learning models, and self-training},
  author={Talak, Rajat and Peng, Lisa R and Carlone, Luca},
  journal={IEEE Transactions on Robotics},
  volume={39},
  number={4},
  pages={2805--2824},
  year={2023}
}

@article{he_tro2023,
  title={ContourPose: Monocular 6-D pose estimation method for reflective textureless metal parts},
  author={He, Zaixing and Li, Quanzhi and Zhao, Xinyue and Wang, Jin and Shen, Huarong and Zhang, Shuyou and Tan, Jianrong},
  journal={IEEE Transactions on Robotics},
  volume={39},
  number={5},
  pages={4037--4050},
  year={2023}
}

@article{shi_tro2023,
  title={Optimal and robust category-level perception: Object pose and shape estimation from 2-D and 3-D semantic keypoints},
  author={Shi, Jingnan and Yang, Heng and Carlone, Luca},
  journal={IEEE Transactions on Robotics},
  volume={39},
  number={5},
  pages={4131--4151},
  year={2023}
}

@article{deng2021poserbpf,
  title={PoseRBPF: A Rao--Blackwellized particle filter for 6-D object pose tracking},
  author={Deng, Xinke and Mousavian, Arsalan and Xiang, Yu and Xia, Fei and Bretl, Timothy and Fox, Dieter},
  journal={IEEE Transactions on Robotics},
  volume={37},
  number={5},
  pages={1328--1342},
  year={2021}
}

@article{cyber2025,
  author={Zhou, Jiaming and Zhu, Qing and Wang, Yaonan and Feng, Mingtao and Liu, Jian and Huang, Jianan and Mian, Ajmal},
  journal={IEEE Transactions on Cybernetics}, 
  title={A state space model for multiobject full 3-d information estimation from rgb-d images}, 
  year={2025},
  volume={55},
  number={5},
  pages={2248-2260}
}

@inproceedings{3dahv,
  title={3d-aware hypothesis \& verification for generalizable relative object pose estimation},
  author={Zhao, Chen and Zhang, Tong and Salzmann, Mathieu},
  booktitle={International Conference on Learning Representations},
  year={2023}
}

@inproceedings{dvmnet,
  title={Dvmnet: Computing relative pose for unseen objects beyond hypotheses},
  author={Zhao, Chen and Zhang, Tong and Dang, Zheng and Salzmann, Mathieu},
  booktitle={Proceedings of the IEEE/CVF Conference on Computer Vision and Pattern Recognition},
  pages={20485--20495},
  year={2024}
}

@inproceedings{one2any,
  title={One2Any: One-reference 6d pose estimation for any object},
  author={Liu, Mengya and Li, Siyuan and Chhatkuli, Ajad and Truong, Prune and Van Gool, Luc and Tombari, Federico},
  booktitle={Proceedings of the Computer Vision and Pattern Recognition Conference},
  pages={6457--6467},
  year={2025}
}

@inproceedings{any6d,
  title={Any6D: Model-free 6d pose estimation of novel objects},
  author={Lee, Taeyeop and Wen, Bowen and Kang, Minjun and Kang, Gyuree and Kweon, In So and Yoon, Kuk-Jin},
  booktitle={Proceedings of the Computer Vision and Pattern Recognition Conference},
  pages={11633--11643},
  year={2025}
}

@ARTICLE{mh6d,
  author={Liu, Jian and Sun, Wei and Liu, Chongpei and Yang, Hui and Zhang, Xing and Mian, Ajmal},
  journal={IEEE Transactions on Neural Networks and Learning Systems}, 
  title={Mh6d: Multi-hypothesis consistency learning for category-level 6-d object pose estimation}, 
  year={2025},
  volume={36},
  number={3},
  pages={4820-4833}
}

@InProceedings{monodiff9d,
  title={Monodiff9d: Monocular category-level 9d object pose estimation via diffusion model},
  author={Liu, Jian and Sun, Wei and Yang, Hui and Zheng, Jin and Geng, Zichen and Rahmani, Hossein and Mian, Ajmal},
  booktitle = {IEEE International Conference on Robotics and Automation},
  year={2025}
}

@inproceedings{li2022dcl,
  title={Dcl-net: Deep correspondence learning network for 6d pose estimation},
  author={Li, Hongyang and Lin, Jiehong and Jia, Kui},
  booktitle={European Conference on Computer Vision},
  pages={369--385},
  year={2022}
}

@inproceedings{cao2022dgecn,
  title={Dgecn: A depth-guided edge convolutional network for end-to-end 6D pose estimation},
  author={Cao, Tuo and Luo, Fei and Fu, Yanping and Zhang, Wenxiao and Zheng, Shengjie and Xiao, Chunxia},
  booktitle={Proceedings of the IEEE/CVF Conference on Computer Vision and Pattern Recognition},
  pages={3783--3792},
  year={2022}
}

@inproceedings{wu2022vote,
  title={Vote from the center: 6 dof pose estimation in rgb-d images by radial keypoint voting},
  author={Wu, Yangzheng and Zand, Mohsen and Etemad, Ali and Greenspan, Michael},
  booktitle={European Conference on Computer Vision},
  pages={335--352},
  year={2022}
}

@inproceedings{zhou2023deep,
  title={Deep fusion transformer network with weighted vector-wise keypoints voting for robust 6d object pose estimation},
  author={Zhou, Jun and Chen, Kai and Xu, Linlin and Dou, Qi and Qin, Jing},
  booktitle={Proceedings of the IEEE/CVF International Conference on Computer Vision},
  pages={13967--13977},
  year={2023}
}

@article{liu2024survey,
  title={Deep learning-based object pose estimation: A comprehensive survey},
  author={Liu, Jian and Sun, Wei and Yang, Hui and Zeng, Zhiwen and Liu, Chongpei and Zheng, Jin and Liu, Xingyu and Rahmani, Hossein and Sebe, Nicu and Mian, Ajmal},  
  journal={International Journal of Computer Vision},
  year={2026},
  volume={134},
  number={81},
  pages={1-45}
}

@inproceedings{chen2022sim,
  title={Sim-to-real 6d object pose estimation via iterative self-training for robotic bin picking},
  author={Chen, Kai and Cao, Rui and James, Stephen and Li, Yichuan and Liu, Yun-Hui and Abbeel, Pieter and Dou, Qi},
  booktitle={European Conference on Computer Vision},
  pages={533--550},
  year={2022}
}

@inproceedings{fu20226d,
  title={6d robotic assembly based on rgb-only object pose estimation},
  author={Fu, Bowen and Leong, Sek Kun and Lian, Xiaocong and Ji, Xiangyang},
  booktitle={IEEE/RSJ International Conference on Intelligent Robots and Systems},
  pages={4736--4742},
  year={2022}
}

@ARTICLE{stg6d,
  author={Liu, Jian and Sun, Wei and Liu, Chongpei and Zhang, Xing and Fu, Qiang},
  journal={IEEE Transactions on Industrial Informatics}, 
  title={Robotic continuous grasping system by shape transformer-guided multiobject category-level 6-d pose estimation}, 
  year={2023},
  volume={19},
  number={11},
  pages={11171-11181}
}

@inproceedings{tyree2022hope,
  title={6-dof pose estimation of household objects for robotic manipulation: An accessible dataset and benchmark},
  author={Tyree, Stephen and Tremblay, Jonathan and To, Thang and Cheng, Jia and Mosier, Terry and Smith, Jeffrey and Birchfield, Stan},
  booktitle={IEEE/RSJ International Conference on Intelligent Robots and Systems},
  pages={13081--13088},
  year={2022}
}

@inproceedings{self6d,
  title={Self6d: Self-supervised monocular 6d object pose estimation},
  author={Wang, Gu and Manhardt, Fabian and Shao, Jianzhun and Ji, Xiangyang and Navab, Nassir and Tombari, Federico},
  booktitle={European Conference on Computer Vision},
  pages={108--125},
  year={2020}
}

@inproceedings{wang2021gdr,
  title={Gdr-net: Geometry-guided direct regression network for monocular 6d object pose estimation},
  author={Wang, Gu and Manhardt, Fabian and Tombari, Federico and Ji, Xiangyang},
  booktitle={Proceedings of the IEEE/CVF Conference on Computer Vision and Pattern Recognition},
  pages={16611--16621},
  year={2021}
}

@inproceedings{di2021so,
  title={So-pose: Exploiting self-occlusion for direct 6d pose estimation},
  author={Di, Yan and Manhardt, Fabian and Wang, Gu and Ji, Xiangyang and Navab, Nassir and Tombari, Federico},
  booktitle={Proceedings of the IEEE/CVF International Conference on Computer Vision},
  pages={12396--12405},
  year={2021}
}

@inproceedings{pvnet,
  title={Pvnet: Pixel-wise voting network for 6dof pose estimation},
  author={Peng, Sida and Liu, Yuan and Huang, Qixing and Zhou, Xiaowei and Bao, Hujun},
  booktitle={Proceedings of the IEEE/CVF Conference on Computer Vision and Pattern Recognition},
  pages={4561--4570},
  year={2019}
}

@inproceedings{ffb6d,
  title={Ffb6d: A full flow bidirectional fusion network for 6d pose estimation},
  author={He, Yisheng and Huang, Haibin and Fan, Haoqiang and Chen, Qifeng and Sun, Jian},
  booktitle={Proceedings of the IEEE/CVF Conference on Computer Vision and Pattern Recognition},
  pages={3003--3013},
  year={2021}
}

@inproceedings{xu20246d,
  title={6d-diff: A keypoint diffusion framework for 6d object pose estimation},
  author={Xu, Li and Qu, Haoxuan and Cai, Yujun and Liu, Jun},
  booktitle={Proceedings of the IEEE/CVF Conference on Computer Vision and Pattern Recognition},
  pages={9676--9686},
  year={2024}
}

@inproceedings{NOCS,
  title={Normalized object coordinate space for category-level 6d object pose and size estimation},
  author={Wang, He and Sridhar, Srinath and Huang, Jingwei and Valentin, Julien and Song, Shuran and Guibas, Leonidas J},
  booktitle={Proceedings of the IEEE/CVF Conference on Computer Vision and Pattern Recognition},
  pages={2642--2651},
  year={2019}
}

@inproceedings{catre,
  title={Catre: Iterative point clouds alignment for category-level object pose refinement},
  author={Liu, Xingyu and Wang, Gu and Li, Yi and Ji, Xiangyang},
  booktitle={European Conference on Computer Vision},
  pages={499--516},
  year={2022}
}

@inproceedings{omni6d,
  title={Omni6d: Large-vocabulary 3d object dataset for category-level 6d object pose estimation},
  author={Zhang, Mengchen and Wu, Tong and Wang, Tai and Wang, Tengfei and Liu, Ziwei and Lin, Dahua},
  booktitle={European Conference on Computer Vision},
  pages={216--232},
  year={2024}
}

@inproceedings{ominnocs,
  title={Ominnocs: A unified nocs dataset and model for 3d lifting of 2d objects},
  author={Krishnan, Akshay and Kundu, Abhijit and Maninis, Kevis-Kokitsi and Hays, James and Brown, Matthew},
  booktitle={European Conference on Computer Vision},
  pages={127--145},
  year={2024}
}

@inproceedings{ttacope,
  title={Tta-cope: Test-time adaptation for category-level object pose estimation},
  author={Lee, Taeyeop and Tremblay, Jonathan and Blukis, Valts and Wen, Bowen and Lee, Byeong-Uk and Shin, Inkyu and Birchfield, Stan and Kweon, In So and Yoon, Kuk-Jin},
  booktitle={Proceedings of the IEEE/CVF Conference on Computer Vision and Pattern Recognition},
  pages={21285--21295},
  year={2023}
}

@inproceedings{georef,
  title={Georef: Geometric alignment across shape variation for category-level object pose refinement},
  author={Zheng, Linfang and Tse, Tze Ho Elden and Wang, Chen and Sun, Yinghan and Chen, Hua and Leonardis, Ales and Zhang, Wei and Chang, Hyung Jin},
  booktitle={Proceedings of the IEEE/CVF Conference on Computer Vision and Pattern Recognition},
  pages={10693--10703},
  year={2024}
}

@inproceedings{HouseCat6D,
  title={Housecat6d--A large-scale multi-modal category level 6d object pose dataset with household objects in realistic scenarios},
  author={Jung, HyunJun and Wu, Shun-Cheng and Ruhkamp, Patrick and others},
  booktitle={Proceedings of the IEEE/CVF Conference on Computer Vision and Pattern Recognition},
  pages={22498--22508},
  year={2024}
}

@article{gou2022unseen,
  title={Unseen object 6D pose estimation: A benchmark and baselines},
  author={Gou, Minghao and Pan, Haolin and Fang, Hao-Shu and Liu, Ziyuan and Lu, Cewu and Tan, Ping},
  journal={arXiv preprint arXiv:2206.11808},
  year={2022}
}

@inproceedings{hagelskjaer2023keymatchnet,
  title={Keymatchnet: Zero-shot pose estimation in 3d point clouds by generalized keypoint matching},
  author={Hagelskj{\ae}r, Frederik and Haugaard, Rasmus Laurvig},
  booktitle={IEEE International Conference on Automation Science and Engineering},
  pages={870--877},
  year={2024}
}

@inproceedings{shugurov2022osop,
  title={Osop: A multi-stage one shot object pose estimation framework},
  author={Shugurov, Ivan and Li, Fu and Busam, Benjamin and Ilic, Slobodan},
  booktitle={Proceedings of the IEEE/CVF Conference on Computer Vision and Pattern Recognition},
  pages={6835--6844},
  year={2022}
}

@inproceedings{nguyen2022templates,
  title={Templates for 3d object pose estimation revisited: Generalization to new objects and robustness to occlusions},
  author={Nguyen, Van Nguyen and Hu, Yinlin and Xiao, Yang and Salzmann, Mathieu and Lepetit, Vincent},
  booktitle={Proceedings of the IEEE/CVF Conference on Computer Vision and Pattern Recognition},
  pages={6771--6780},
  year={2022}
}

@inproceedings{fan2024pope,
  title={POPE: 6-DoF Promptable Pose Estimation of Any Object in Any Scene with One Reference},
  author={Fan, Zhiwen and Pan, Panwang and Wang, Peihao and Jiang, Yifan and Xu, Dejia and Wang, Zhangyang},
  booktitle={Proceedings of the IEEE/CVF Conference on Computer Vision and Pattern Recognition},
  pages={7771--7781},
  year={2024}
}

@inproceedings{zhao2024locposenet,
  title={Locposenet: Robust location prior for unseen object pose estimation},
  author={Zhao, Chen and Hu, Yinlin and Salzmann, Mathieu},
  booktitle={International Conference on 3D Vision},
  pages={1072--1081},
  year={2024}
}

@inproceedings{pan2024learning,
  title={Learning to estimate 6dof pose from limited data: A few-shot, generalizable approach using rgb images},
  author={Pan, Panwang and Fan, Zhiwen and Feng, Brandon Y and Wang, Peihao and Li, Chenxin and Wang, Zhangyang},
  booktitle={International Conference on 3D Vision},
  pages={1059--1071},
  year={2024}
}

@inproceedings{pitteri2019cornet,
  title={CorNet: Generic 3d corners for 6d pose estimation of new objects without retraining},
  author={Pitteri, Giorgia and Ilic, Slobodan and Lepetit, Vincent},
  booktitle={Proceedings of the IEEE/CVF International Conference on Computer Vision Workshops},
  year={2019}
}

@inproceedings{pitteri20203d,
  title={3d object detection and pose estimation of unseen objects in color images with local surface embeddings},
  author={Pitteri, Giorgia and Bugeau, Aur{\'e}lie and Ilic, Slobodan and Lepetit, Vincent},
  booktitle={Proceedings of the Asian Conference on Computer Vision},
  year={2020}
}

@inproceedings{sundermeyer2020multi,
  title={Multi-path learning for object pose estimation across domains},
  author={Sundermeyer, Martin and Durner, Maximilian and Puang, En Yen and Marton, Zoltan-Csaba and Vaskevicius, Narunas and Arras, Kai O and Triebel, Rudolph},
  booktitle={Proceedings of the IEEE/CVF Conference on Computer Vision and Pattern Recognition},
  pages={13916--13925},
  year={2020}
}

@inproceedings{okorn2021zephyr,
  title={Zephyr: Zero-shot pose hypothesis rating},
  author={Okorn, Brian and Gu, Qiao and Hebert, Martial and Held, David},
  booktitle={IEEE International Conference on Robotics and Automation},
  pages={14141--14148},
  year={2021}
}

@inproceedings{corsetti2024open,
  title={Open-vocabulary object 6D pose estimation},
  author={Corsetti, Jaime and Boscaini, Davide and Oh, Changjae and Cavallaro, Andrea and Poiesi, Fabio},
  booktitle={Proceedings of the IEEE/CVF Conference on Computer Vision and Pattern Recognition},
  pages={18071--18080},
  year={2024}
}

@inproceedings{wu2021unseen,
  title={Unseen object pose estimation via registration},
  author={Wu, Jun and Wang, Yue and Xiong, Rong},
  booktitle={IEEE International Conference on Real-time Computing and Robotics},
  pages={974--979},
  year={2021}
}

@inproceedings{bundlesdf,
  title={Bundlesdf: Neural 6-dof tracking and 3d reconstruction of unknown objects},
  author={Wen, Bowen and Tremblay, Jonathan and Blukis, Valts and Tyree, Stephen and M{\"u}ller, Thomas and Evans, Alex and Fox, Dieter and Kautz, Jan and Birchfield, Stan},
  booktitle={Proceedings of the IEEE/CVF Conference on Computer Vision and Pattern Recognition},
  pages={606--617},
  year={2023}
}

@inproceedings{nope,
  title={Nope: Novel object pose estimation from a single image},
  author={Nguyen, Van Nguyen and Groueix, Thibault and Ponimatkin, Georgy and Hu, Yinlin and Marlet, Renaud and Salzmann, Mathieu and Lepetit, Vincent},
  booktitle={Proceedings of the IEEE/CVF Conference on Computer Vision and Pattern Recognition},
  pages={17923--17932},
  year={2024}
}

@article{Dang2024Match,
  author={Dang, Zheng and Wang, Lizhou and Guo, Yu and Salzmann, Mathieu},
  journal={IEEE Transactions on Pattern Analysis and Machine Intelligence}, 
  volume={46},
  number={06},
  pages={4489--4503},
  title={Match normalization: Learning-based point cloud registration for 6d object pose estimation in the real world}, 
  year={2024}
}

@inproceedings{GeoTransformer,
  title={Geometric transformer for fast and robust point cloud registration},
  author={Qin, Zheng and Yu, Hao and Wang, Changjian and Guo, Yulan and Peng, Yuxing and Xu, Kai},
  booktitle={Proceedings of the IEEE/CVF Conference on Computer Vision and Pattern Recognition},
  pages={11143--11152},
  year={2022}
}

@article{chen2023zeropose,
  title={Zeropose: Cad-model-based zero-shot pose estimation},
  author={Chen, Jianqiu and Sun, Mingshan and Bao, Tianpeng and Zhao, Rui and Wu, Liwei and He, Zhenyu},
  journal={arXiv preprint arXiv:2305.17934},
  year={2023}
}

@inproceedings{GCPose,
  title={Learning symmetry-aware geometry correspondences for 6d object pose estimation},
  author={Zhao, Heng and Wei, Shenxing and Shi, Dahu and Tan, Wenming and Li, Zheyang and Ren, Ye and Wei, Xing and Yang, Yi and Pu, Shiliang},
  booktitle={Proceedings of the IEEE/CVF International Conference on Computer Vision},
  pages={14045--14054},
  year={2023}
}

@inproceedings{SAM-6D,
  title={Sam-6d: Segment anything model meets zero-shot 6d object pose estimation},
  author={Lin, Jiehong and Liu, Lihua and Lu, Dekun and Jia, Kui},
  booktitle={Proceedings of the IEEE/CVF Conference on Computer Vision and Pattern Recognition},
  pages={27906--27916},
  year={2024}
}

@inproceedings{MatchU,
  title={Matchu: Matching unseen objects for 6d pose estimation from rgb-d images},
  author={Huang, Junwen and Yu, Hao and Yu, Kuan-Ting and Navab, Nassir and Ilic, Slobodan and Busam, Benjamin},
  booktitle={Proceedings of the IEEE/CVF Conference on Computer Vision and Pattern Recognition},
  pages={10095--10105},
  year={2024}
}

@inproceedings{FreeZe,
  author = {Caraffa, Andrea and Boscaini, Davide and Hamza, Amir and Poiesi, Fabio},
  title = {FreeZe: Training-free zero-shot 6D pose estimation with geometric and vision foundation models},
  booktitle={European Conference on Computer Vision},
  year = {2024},
}

@inproceedings{foundpose,
  title={Foundpose: Unseen object pose estimation with foundation features},
  author={{\"O}rnek, Evin P{\i}nar and Labb{\'e}, Yann and Tekin, Bugra and Ma, Lingni and Keskin, Cem and Forster, Christian and Hodan, Tomas},
  booktitle={European Conference on Computer Vision},
  pages={163--182},
  year={2024}
}

@inproceedings{wang2024object,
  title={Object pose estimation via the aggregation of diffusion features},
  author={Wang, Tianfu and Hu, Guosheng and Wang, Hongguang},
  booktitle={Proceedings of the IEEE/CVF Conference on Computer Vision and Pattern Recognition},
  pages={10238--10247},
  year={2024}
}

@inproceedings{megapose,
  title={Megapose: 6d pose estimation of novel objects via render \& compare},
  author={Labb{\'e}, Yann and Manuelli, Lucas and Mousavian, Arsalan and Tyree, Stephen and Birchfield, Stan and Tremblay, Jonathan and Carpentier, Justin and Aubry, Mathieu and Fox, Dieter and Sivic, Josef},
  booktitle={Proceedings of the 6th Conference on Robot Learning},
  year={2022}
}

@inproceedings{genflow,
  title={Genflow: Generalizable recurrent flow for 6d pose refinement of novel objects},
  author={Moon, Sungphill and Son, Hyeontae and Hur, Dongcheol and Kim, Sangwook},
  booktitle={Proceedings of the IEEE/CVF Conference on Computer Vision and Pattern Recognition},
  pages={10039--10049},
  year={2024}
}

@inproceedings{gigapose,
  title={Gigapose: Fast and robust novel object pose estimation via one correspondence},
  author={Nguyen, Van Nguyen and Groueix, Thibault and Salzmann, Mathieu and Lepetit, Vincent},
  booktitle={Proceedings of the IEEE/CVF Conference on Computer Vision and Pattern Recognition},
  pages={9903--9913},
  year={2024}
}

@inproceedings{foundationpose,
  title={Foundationpose: Unified 6d pose estimation and tracking of novel objects},
  author={Wen, Bowen and Yang, Wei and Kautz, Jan and Birchfield, Stan},
  booktitle={Proceedings of the IEEE/CVF Conference on Computer Vision and Pattern Recognition},
  pages={17868--17879},
  year={2024}
}

@inproceedings{onepose,
  title={Onepose: One-shot object pose estimation without cad models},
  author={Sun, Jiaming and Wang, Zihao and Zhang, Siyu and He, Xingyi and Zhao, Hongcheng and Zhang, Guofeng and Zhou, Xiaowei},
  booktitle={Proceedings of the IEEE/CVF Conference on Computer Vision and Pattern Recognition},
  pages={6825--6834},
  year={2022}
}

@inproceedings{onepose++,
  title={Onepose++: Keypoint-free one-shot object pose estimation without CAD models},
  author={He, Xingyi and Sun, Jiaming and Wang, Yuang and Huang, Di and Bao, Hujun and Zhou, Xiaowei},
  booktitle={Advances in Neural Information Processing Systems},
  volume={35},
  pages={35103--35115},
  year={2022}
}

@inproceedings{fs6d,
  title={Fs6d: Few-shot 6d pose estimation of novel objects},
  author={He, Yisheng and Wang, Yao and Fan, Haoqiang and Sun, Jian and Chen, Qifeng},
  booktitle={Proceedings of the IEEE/CVF Conference on Computer Vision and Pattern Recognition},
  pages={6814--6824},
  year={2022}
}

@inproceedings{castro2023posematcher,
  title={Posematcher: One-shot 6d object pose estimation by deep feature matching},
  author={Castro, Pedro and Kim, Tae-Kyun},
  booktitle={Proceedings of the IEEE/CVF International Conference on Computer Vision},
  pages={2148--2157},
  year={2023}
}

@inproceedings{lee2024mfos,
  title={Mfos: Model-free \& one-shot object pose estimation},
  author={Lee, JongMin and Cabon, Yohann and Br{\'e}gier, Romain and Yoo, Sungjoo and Revaud, Jerome},
  booktitle={Proceedings of the AAAI Conference on Artificial Intelligence},
  volume={38},
  number={4},
  pages={2911--2919},
  year={2024}
}

@inproceedings{sun2021loftr,
  title={LoFTR: Detector-free local feature matching with transformers},
  author={Sun, Jiaming and Shen, Zehong and Wang, Yuang and Bao, Hujun and Zhou, Xiaowei},
  booktitle={Proceedings of the IEEE/CVF Conference on Computer Vision and Pattern Recognition},
  pages={8922--8931},
  year={2021}
}

@inproceedings{gen6d,
  title={Gen6d: Generalizable model-free 6-dof object pose estimation from rgb images},
  author={Liu, Yuan and Wen, Yilin and Peng, Sida and Lin, Cheng and Long, Xiaoxiao and Komura, Taku and Wang, Wenping},
  booktitle={European Conference on Computer Vision},
  pages={298--315},
  year={2022}
}

@inproceedings{latentfusion,
  title={Latentfusion: End-to-end differentiable reconstruction and rendering for unseen object pose estimation},
  author={Park, Keunhong and Mousavian, Arsalan and Xiang, Yu and Fox, Dieter},
  booktitle={Proceedings of the IEEE/CVF Conference on Computer Vision and Pattern Recognition},
  pages={10710--10719},
  year={2020}
}

@inproceedings{du2022pizza,
  title={Pizza: A powerful image-only zero-shot zero-cad approach to 6 dof tracking},
  author={Du, Yuming and Xiao, Yang and Ramamonjisoa, Michael and Lepetit, Vincent and others},
  booktitle={International Conference on 3D Vision},
  pages={515--525},
  year={2022}
}

@inproceedings{sa6d,
  title={Sa6d: Self-adaptive few-shot 6d pose estimator for novel and occluded objects},
  author={Gao, Ning and Ngo, Vien Anh and Ziesche, Hanna and Neumann, Gerhard},
  booktitle={7th Annual Conference on Robot Learning},
  year={2023}
}

@inproceedings{cai2024gs,
  title={Gs-pose: Cascaded framework for generalizable segmentation-based 6d object pose estimation},
  author={Cai, Dingding and Heikkil{\"a}, Janne and Rahtu, Esa},
  booktitle={International Conference on 3D Vision},
  year={2025}
}

@inproceedings{linemod,
  title={Multimodal templates for real-time detection of texture-less objects in heavily cluttered scenes},
  author={Hinterstoisser, Stefan and Holzer, Stefan and Cagniart, Cedric and Ilic, Slobodan and Konolige, Kurt and Navab, Nassir and Lepetit, Vincent},
  booktitle={Proceedings of the IEEE/CVF International Conference on Computer Vision},
  pages={858--865},
  year={2011}
}

@inproceedings{linemod-o,
  title={Learning 6d object pose estimation using 3d object coordinates},
  author={Brachmann, Eric and Krull, Alexander and Michel, Frank and Gumhold, Stefan and Shotton, Jamie and Rother, Carsten},
  booktitle={European Conference on Computer Vision},
  pages={536--551},
  year={2014}
}

@inproceedings{2024bop,
  title={Bop challenge 2023 on detection segmentation and pose estimation of seen and unseen rigid objects},
  author={Hodan, Tomas and Sundermeyer, Martin and Labbe, Yann and Nguyen, Van Nguyen and Wang, Gu and Brachmann, Eric and Drost, Bertram and Lepetit, Vincent and Rother, Carsten and Matas, Jiri},
  booktitle={Proceedings of the IEEE/CVF Conference on Computer Vision and Pattern Recognition},
  pages={5610--5619},
  year={2024}
}

@inproceedings{IC-BIN,
  title={Recovering 6d object pose and predicting next-best-view in the crowd},
  author={Doumanoglou, Andreas and Kouskouridas, Rigas and Malassiotis, Sotiris and Kim, Tae-Kyun},
  booktitle={Proceedings of the IEEE/CVF Conference on Computer Vision and Pattern Recognition},
  pages={3583--3592},
  year={2016}
}

@inproceedings{HB,
  title={Homebreweddb: Rgb-d dataset for 6d pose estimation of 3d objects},
  author={Kaskman, Roman and Zakharov, Sergey and Shugurov, Ivan and Ilic, Slobodan},
  booktitle={Proceedings of the IEEE/CVF International Conference on Computer Vision Workshops},
  year={2019}
}

@inproceedings{tless,
  title={T-less: An rgb-d dataset for 6d pose estimation of texture-less objects},
  author={Hodan, Tom{\'a}{\v{s}} and Haluza, Pavel and Obdr{\v{z}}{\'a}lek, {\v{S}}tep{\'a}n and Matas, Jiri and Lourakis, Manolis and Zabulis, Xenophon},
  booktitle={Proceedings of the IEEE/CVF Winter Conference on Applications of Computer Vision},
  pages={880--888},
  year={2017}
}

@inproceedings{itodd,
  title={Introducing mvtec itodd-a dataset for 3d object recognition in industry},
  author={Drost, Bertram and Ulrich, Markus and Bergmann, Paul and Hartinger, Philipp and Steger, Carsten},
  booktitle={Proceedings of the IEEE/CVF International Conference on Computer Vision Workshops},
  pages={2200--2208}
}

@inproceedings{posecnn,
  title={Posecnn: A convolutional neural network for 6d object pose estimation in cluttered scenes},
  author={Xiang, Yu and Schmidt, Tanner and Narayanan, Venkatraman and Fox, Dieter},
  booktitle={Robotics: Science and Systems},
  year={2018}
}

@article{shapenet,
  title={Shapenet: An information-rich 3d model repository},
  author={Chang, Angel X and Funkhouser, Thomas and Guibas, Leonidas and others},
  journal={arXiv preprint arXiv:1512.03012},
  year={2015}
}

@inproceedings{googlescan,
  title={Google scanned objects: A high-quality dataset of 3d scanned household items},
  author={Downs, Laura and Francis, Anthony and Koenig, Nate and Kinman, Brandon and Hickman, Ryan and Reymann, Krista and McHugh, Thomas B and Vanhoucke, Vincent},
  booktitle={IEEE International Conference on Robotics and Automation},
  pages={2553--2560},
  year={2022}
}

@inproceedings{add,
  title={Model based training, detection and pose estimation of texture-less 3d objects in heavily cluttered scenes},
  author={Hinterstoisser, Stefan and Lepetit, Vincent and Ilic, Slobodan and Holzer, Stefan and Bradski, Gary and Konolige, Kurt and Navab, Nassir},
  booktitle={Asian Conference on Computer Vision},
  pages={548--562},
  year={2013}
}

@inproceedings{pointtransformer,
  title={Point transformer},
  author={Zhao, Hengshuang and Jiang, Li and Jia, Jiaya and Torr, Philip HS and Koltun, Vladlen},
  booktitle={Proceedings of the IEEE/CVF International Conference on Computer Vision},
  pages={16259--16268},
  year={2021}
}

@inproceedings{vit,
  title={An image is worth 16x16 words: Transformers for image recognition at scale},
  author={Dosovitskiy, Alexey and others},
  booktitle={International Conference on Learning Representations},
  year={2021}
}

@inproceedings{maskrcnn,
  title={Mask r-cnn},
  author={He, Kaiming and Gkioxari, Georgia and Doll{\'a}r, Piotr and Girshick, Ross},
  booktitle={Proceedings of the IEEE/CVF International Conference on Computer Vision},
  pages={2961--2969},
  year={2017}
}

@inproceedings{cnos,
  title={Cnos: A strong baseline for cad-based novel object segmentation},
  author={Nguyen, Van Nguyen and Groueix, Thibault and Ponimatkin, Georgy and Lepetit, Vincent and Hodan, Tomas},
  booktitle={Proceedings of the IEEE/CVF International Conference on Computer Vision},
  pages={2134--2140},
  year={2023}
}

@inproceedings{S6model,
  title={Mamba: Linear-time sequence modeling with selective state spaces},
  author={Gu, Albert and Dao, Tri},
  booktitle={Conference on Language Modeling},
  year={2024}
}

@inproceedings{S4model,
  title={Efficiently modeling long sequences with structured state spaces},
  author={Gu, Albert and Goel, Karan and R{\'e}, Christopher},
  booktitle={International Conference on Learning Representations},
  year={2022}
}

@inproceedings{Vmamba,
  title={Vmamba: Visual state space model},
  author={Liu, Yue and Tian, Yunjie and Zhao, Yuzhong and Yu, Hongtian and Xie, Lingxi and Wang, Yaowei and Ye, Qixiang and Jiao, Jianbin and Liu, Yunfan},
  booktitle={Advances in Neural Information Processing Systems},
  year={2024}
}

@inproceedings{unopose,
  title={Unopose: Unseen object pose estimation with an unposed rgb-d reference image},
  author={Liu, Xingyu and Wang, Gu and Zhang, Ruida and Zhang, Chenyangguang and Tombari, Federico and Ji, Xiangyang},
  booktitle={Proceedings of the Computer Vision and Pattern Recognition Conference},
  pages={22023--22034},
  year={2025}
}

@article{Diff9D,
  author={Liu, Jian and Sun, Wei and Yang, Hui and Deng, Pengchao and Liu, Chongpei and Sebe, Nicu and Rahmani, Hossein and Mian, Ajmal},
  title={Diff9d: Diffusion-based domain-generalized category-level 9-dof object pose estimation},
  journal={IEEE Transactions on Pattern Analysis and Machine Intelligence},
  year={2025},
  volume={47},
  number={7},
  pages={5520-5537}
}

@inproceedings{ist-net,
  title={Ist-net: Prior-free category-level pose estimation with implicit space transformation},
  author={Liu, Jianhui and Chen, Yukang and Ye, Xiaoqing and Qi, Xiaojuan},
  booktitle={Proceedings of the IEEE/CVF International Conference on Computer Vision},
  pages={13978--13988},
  year={2023}
}

@inproceedings{gpv-pose,
  title={Gpv-pose: Category-level object pose estimation via geometry-guided point-wise voting},
  author={Di, Yan and Zhang, Ruida and Lou, Zhiqiang and Manhardt, Fabian and Ji, Xiangyang and Navab, Nassir and Tombari, Federico},
  booktitle={Proceedings of the IEEE/CVF Conference on Computer Vision and Pattern Recognition},
  pages={6781--6791},
  year={2022}
}

@article{dinov2,
  title={Dinov2: Learning robust visual features without supervision},
  author={Oquab, Maxime and Darcet, Timoth{\'e}e and Moutakanni, Th{\'e}o and Vo, Huy and Szafraniec, Marc and Khalidov, Vasil and Fernandez, Pierre and Haziza, Daniel and Massa, Francisco and El-Nouby, Alaaeldin and others},
  journal={Transactions on Machine Learning Research},
  year={2024}
}

@inproceedings{predator,
  title={Predator: Registration of 3d point clouds with low overlap},
  author={Huang, Shengyu and Gojcic, Zan and Usvyatsov, Mikhail and Wieser, Andreas and Schindler, Konrad},
  booktitle={Proceedings of the IEEE/CVF Conference on Computer Vision and Pattern Recognition},
  pages={4267--4276},
  year={2021}
}

\end{document}